\documentclass[nohyperref]{article}

\usepackage{microtype}
\usepackage{graphicx}
\usepackage{subfigure}
\usepackage{booktabs} 

\usepackage{hyperref}

\usepackage[accepted]{icml2022}

\usepackage{amsmath}
\usepackage{amssymb}
\usepackage{mathtools}
\usepackage{amsthm}

\usepackage{enumitem}

\usepackage[colorinlistoftodos,prependcaption,textsize=tiny]{todonotes}

\usepackage{amsmath}

  \def\mC{{\mathcal C}}
  \def\mD{{\mathcal D}}

  \def\mL{{\mathcal L}}

  \def\mR{{\mathcal R}}

  \def\mX{{\mathcal X}}

  \DeclareMathAlphabet\mathbfcal{OMS}{cmsy}{b}{n}

  \def\0{{\bf 0}}
  \def\1{{\bf 1}}

  \def\bmm{{\bf m}}

  \def\bx{{\bf x}}
  \def\by{{\bf y}}

  \def\bx{{\bf x}}
  
  \def\by{{\bf y}}

\def\eg{\emph{e.g.}} 
\def\ie{\emph{i.e.}} 
 
 \def\vs{\emph{vs.}}
\def\wrt{{w.r.t.~}}

\usepackage[capitalize,noabbrev]{cleveref}

\def\camera{\textcolor{black}}

\usepackage{balance}
\usepackage{xspace}
\usepackage{pifont}

\newcommand{\myeata}{EATA\xspace}
\newcommand{\myeta}{ETA\xspace}

\definecolor{chengreen}{RGB}{17, 120, 100}

\usepackage{threeparttable}
\usepackage{multicol,multirow}
\usepackage{booktabs} 
\usepackage{bbding}

\theoremstyle{plain}

\theoremstyle{definition}

\theoremstyle{remark}

\usepackage[textsize=tiny]{todonotes}

\def\mytitle{Efficient Test-Time Model Adaptation without Forgetting}

\icmltitlerunning{\mytitle}

\begin{document}

\twocolumn[
\icmltitle{\mytitle}



\icmlsetsymbol{equal}{*}

\begin{icmlauthorlist}
\icmlauthor{Shuaicheng Niu}{equal,scut,edu}
\icmlauthor{Jiaxiang Wu}{equal,tencent}
\icmlauthor{Yifan Zhang}{equal,nus} 
\icmlauthor{Yaofo Chen}{scut} \\
\icmlauthor{Shijian Zheng}{scut}
\icmlauthor{Peilin Zhao}{tencent}
\icmlauthor{Mingkui Tan}{scut,pz}
\end{icmlauthorlist}

\icmlaffiliation{scut}{School of Software Engineering, South China University of Technology, China}
\icmlaffiliation{pz}{Pazhou Laboratory, China}
\icmlaffiliation{tencent}{Tencent AI Lab, China}
\icmlaffiliation{nus}{National University of Singapore, Singapore}
\icmlaffiliation{edu}{Key Laboratory of Big Data and Intelligent Robot, Ministry of Education, China}

\icmlcorrespondingauthor{Mingkui Tan}{mingkuitan@scut.edu.cn}

\icmlkeywords{Machine Learning, ICML}

\vskip 0.3in
]



\printAffiliationsAndNotice{\icmlEqualContribution} 

\begin{abstract}

Test-time adaptation (TTA) seeks to tackle potential distribution shifts between training and testing data by adapting a given model w.r.t. any testing sample. This task is particularly important for deep models when the test environment changes frequently. Although some recent attempts have been made to handle this task, we still face two practical challenges: 1) existing methods have to perform backward computation for each test sample, resulting in unbearable prediction cost to many applications; 2) while existing TTA solutions can significantly improve the test performance on out-of-distribution data, they often suffer from severe performance degradation on in-distribution data after TTA  (known as catastrophic forgetting). In this paper, we point out that not all the test samples contribute equally to model adaptation, and high-entropy ones may lead to noisy gradients that could disrupt the model. Motivated by this, we propose an active sample selection criterion to identify reliable and non-redundant samples, on which the model is updated to minimize the entropy loss for test-time adaptation. Furthermore, to alleviate the forgetting issue, we introduce a Fisher regularizer to constrain important model parameters from drastic changes, where the Fisher importance is estimated from test samples with generated pseudo labels. Extensive experiments on CIFAR-10-C, ImageNet-C, and ImageNet-R verify the effectiveness of our proposed method. Code is available at \href{https://github.com/mr-eggplant/EATA}{https://github.com/mr-eggplant/EATA}.
\end{abstract}

\section{Introduction}

Deep neural networks (DNNs) have achieved excellent performance in many challenging tasks, including image classification~\cite{he2016deep,tan2019efficientnet}, video recognition~\cite{wang2018nonlocal,chen2021rspnet}, and many other areas~\cite{choi2018stargan,fan2020inf,xu2021towards}. One prerequisite behind the success of DNNs is that the test samples are drawn from the same distribution as the training data, which, however, is often violated in many real-world applications. In practice, test samples may encounter natural variations or corruptions (also called \emph{distribution shift}), such as changes in lighting resulting from weather change and unexpected noises resulting from sensor degradation~\cite{hendrycks2019benchmarking,koh2021wilds}. Unfortunately, models are often very sensitive to such distribution shifts and suffer severe performance degradation. 

Recently, several attempts~\cite{sun2020test,wang2021tent,liu2021ttt++,zhang2021memo,zhang2021test,wang2022continual} have been proposed to handle the distribution shifts by online adapting a model at test time (called \textit{test-time adaptation}). Test-time training (TTT)~\cite{sun2020test} first proposes this pipeline. Given a test sample, TTT first fine-tunes the model via rotation classification~\cite{gidaris2018unsupervised} and then makes a prediction using the updated model. However, TTT still relies on additional training modifications (adding rotation head into the model), and thus the access to original training data is also compulsory. These requirements may be infeasible if, \eg, the training data is unavailable due to privacy/storage concerns or the training involves unexpected heavy computation. To avoid these, Tent~\cite{wang2021tent} and MEMO~\cite{zhang2021memo} propose methods for fully test-time adaptation, in which the adaptation only involves  test samples and a trained model.

\begin{table*}[t!]
\caption{Characteristics of problem settings that adapt a trained model to a potentially shifted test domain. `Offline' adaptation assumes access to the entire source or target dataset, while `Online' adaptation can predict a single or batch of incoming test samples immediately.}
\label{tab:diff-settings}
\newcommand{\tabincell}[2]{\begin{tabular}{@{}#1@{}}#2\end{tabular}}
\begin{center}
\begin{threeparttable}
    \resizebox{1.0\linewidth}{!}{
 	\begin{tabular}{l|cccccc|c}
 	\toprule
 	 Setting & Source Data & Target Data & Training Loss & Testing Loss & Offline & Online & Source Acc.  \\
 	\midrule
        Fine-tuning   & \XSolidBrush & $\bx^t, y^t$ & $\mL(\bx^t,y^t)$ & -- & \Checkmark & \XSolidBrush & Not Considered  \\
        Continual learning   & \XSolidBrush & $\bx^t, y^t$ & $\mL(\bx^t,y^t)$ & -- & \Checkmark & \XSolidBrush & Maintained  \\
        Unsupervised domain adaptation   & $\bx^s,y^s$ & $\bx^t$ & $\mL(\bx^s,y^s)+\mL(\bx^s,\bx^t)$ & -- & \Checkmark & \XSolidBrush & Maintained  \\
        Test-time training   & $\bx^s,y^s$ & $\bx^t$ & $\mL(\bx^s,y^s)+\mL(\bx^s)$ & $\mL(\bx^t)$ & \XSolidBrush & \Checkmark &  Not Considered   \\
        Fully test-time adaptation (FTTA) & \XSolidBrush & $\bx^t$ & \XSolidBrush & $\mL(\bx^t)$ & \XSolidBrush & \Checkmark & Not Considered   \\
        \midrule
        \myeata (ours)   & \XSolidBrush & $\bx^t$ & \XSolidBrush & $\mL(\bx^t)$ & \XSolidBrush & \Checkmark & Maintained   \\
    \bottomrule
	\end{tabular}
	}
\end{threeparttable}
\end{center}
\end{table*}

Although recent test-time adaptation methods are effective at handling test shifts, they still suffer the following limitations. First, since we adapt models at test time, the adaptation efficiency is quite important in many latency-sensitive scenarios. However, prior methods rely on performing backward computation for each test sample (even multiple backward passes for a single sample, such as TTT~\cite{sun2020test} and MEMO~\cite{zhang2021memo}). As performing back-propagation too much is time-consuming, these approaches may be infeasible when the latency is unacceptable. Second, these methods focus on boosting the performance of a trained model on out-of-distribution (OOD) test samples, ignoring that the model after test-time adaptation suffers a severe performance degradation (named \textit{forgetting}) on in-distribution (ID) test samples (see Figure~\ref{fig:imageC-forgetting-level-5}). An eligible test-time adaptation approach should perform well on both OOD and ID test samples simultaneously, since test samples actually often come with both ID and OOD domains.

To address above limitations, we propose an Efficient Anti-forgetting Test-time Adaptation method (namely \myeata), which consists of a sample-efficient optimization strategy and a weight regularizer. Specifically, we devise a sample-adaptive entropy minimization loss, in which we exclude two types of samples out of optimization: i) samples with high entropy values, since the gradients provided by those samples are highly unreliable; and ii) samples are very similar. In this case, the total number of backward updates of a test data streaming is properly reduced (improving efficiency) and the model performance on OOD samples is also improved. On the other hand, we devise an anti-forgetting regularizer   to enforce the important weights of the model do not change a lot during the adaptation. We calculate the weight importance based on Fisher information~\cite{kirkpatrick2017overcoming} via a small set of  test samples. With this regularization, the model can be continually adapted without performance degradation on ID test samples.

\textbf{Contributions:} 1) We propose an active sample identification scheme to filter out non-reliable and redundant test data from model adaptation; 2) We extend the label-dependent Fisher regularizer to test samples with pseudo label generation, which prevents drastic changes in important model weights; and 3) We demonstrate that the proposed \myeata improves the efficiency of test-time adaptation and also alleviates the long-neglected catastrophic forgetting issue.

\begin{figure*}[t]

\centering
\includegraphics[width=0.96\linewidth]{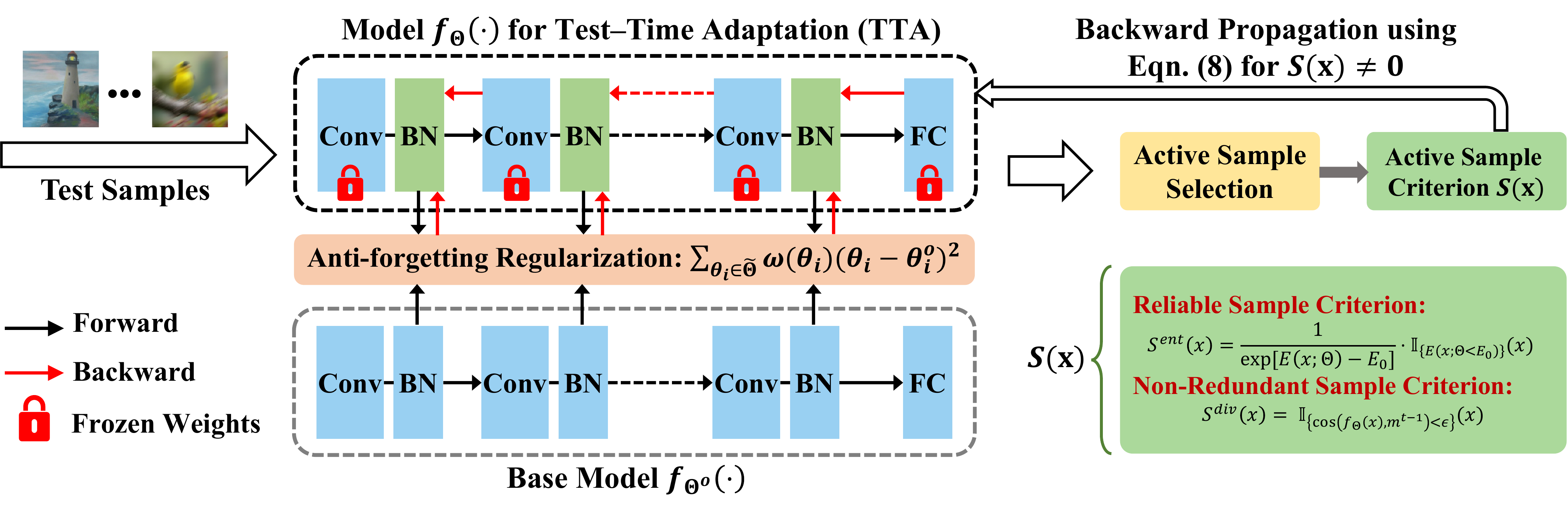}
\vspace{-0.15in}
\caption{An illustration of the proposed \myeata. 
Given a trained base model $f_{\Theta^o}$, we perform test-time adaptation with a model $f_{\Theta}$ that initialized from $\Theta^o$.
During the adaptation process, we only update the parameters of batch normalization layers in $f_{\Theta}$ and froze the rest parameters.
When a batch of test sample $\mX\small{=}\{\bx_b\}_{b=1}^{B}$ come, we calculate a sample-adaptive weight $S(\bx)$ for each test sample to identify whether the sample is active for adaptation or not.
We only perform backward propagation with the samples whose $S(\bx) \neq 0$.
Moreover, we propose an anti-forgetting regularizer to prevent the model parameters $\Theta$ changing too much from $\Theta^o$.}
\label{fig:overall_illustration}
\end{figure*}

\section{Related Work}

We divide the discussion on related works based on the different adaptation settings summarized in Table~\ref{tab:diff-settings}.

\noindent\textbf{Test-Time Adaptation} (TTA) aims to improve model accuracy on OOD test data through model adaptation with  test samples.
Existing test-time training methods, \eg, TTT~\cite{sun2020test} and TTT++~\cite{liu2021ttt++}, jointly train a source model via both supervised and self-supervised objectives, and then adapt the model via self-supervised objective at test time. This pipeline, however, has assumptions on the manner of model training, which may not always be controllable in practice. To address this, fully test-time adaptation methods have been proposed to adapt a model with only  test data, including batchnorm statistics adaptation~\cite{nado2020evaluating,schneider2020improving,khurana2021sita}, test-time entropy minimization~\cite{wang2021tent,fleuret2021test}, prediction consistency maximization over different augmentations~\cite{zhang2021memo}, and classifier adjustment~\cite{iwasawa2021test}. Our work follows the fully test-time adaptation setting and seeks to address two key limitations of prior works (i.e., efficiency hurdle and catastrophic forgetting) to make test-time adaptation more practical in real-world applications.

\noindent\textbf{Continual Learning} (CL) aims to help the model remember the essential concepts that have been learned previously, alleviating the catastrophic forgetting issue when learning on a new task~\cite{kirkpatrick2017overcoming,li2017learning,rolnick2019experience,farajtabar2020orthogonal,niu2021disturbance,Mittal_2021_CVPR}. In our work, we share the same motivation as CL and point out that test-time adaptation also suffers catastrophic forgetting (\ie, performance degradation on ID test samples), which makes test-time adaptation approaches are unstable to deploy. To conquer this, we propose a simple yet effective solution to maintain the model performance on ID test samples (by only using  test data) and meanwhile improve the performance on OOD test samples. 

\noindent\textbf{Unsupervised Domain Adaptation (UDA).} Conventional UDA tackles distribution shifts by jointly optimizing a source model on both labeled source data and unlabeled target data, such as devising a domain discriminator to learn domain-invariant features~\cite{pei2018multi,saito2018maximum,zhang2020collaborative,zhang2020covid}. To avoid access to source data, recent source-free UDA methods are proposed either by generative modeling~\cite{li2020model,kundu2020universal,Qiu2021CPGA} or information maximization~\cite{liang2020we}. Nevertheless, such methods optimize offline via multiple epochs and losses. In contrast, our method adapts in an online manner and selectively performs once backward propagation for one given target sample, which is more efficient during inference. 

\section{Problem Formulation}

Without loss of generality, let $P\left( \bx \right)$ be the distribution of training data $\{\bx_i\}_{i=1}^{N}$  (namely $\bx_i \sim P\left( \bx \right)$) and $f_{\Theta^o}(\bx)$ be a \textbf{base model} trained on labeled training data $\{(\bx_i,y_i)\}_{i=1}^{N}$, where $\Theta^o$ denotes the model parameters. Due to the training process, the model $f_{\Theta^o}(\bx)$ tends to fit (or overfit) the training data.
During the inference state, the model shall perform well for  the in-distribution testing data, namely $\bx \sim P\left( \bx \right)$. However, in practice, due to possible distribution shifts between training and test data, we may encounter many out-of-distribution test samples, namely $\bx \sim Q\left( \bx \right)$, where $Q\left( \bx \right) \neq P\left( \bx \right)$. In this case, the prediction would be very unreliable and the performance is also very poor.  

Test-time adaptation (TTA)~\cite{wang2021tent,zhang2021memo} aims at boosting the  out-of-distribution prediction performance by  doing model adaptation on test data only. Specifically, given a set of test samples $\{\bx_j\}_{j=1}^{M}$, where $\bx_j \sim Q \left( \bx \right)$ and $Q\left( \bx \right) \neq P\left( \bx \right)$, one needs to adapt $f_{\Theta}(\bx)$ to improve the prediction performance with on test data in any cases. To achieve this, existing methods often seek to update the model by minimizing some unsupervised objective defined on test samples:
\begin{equation}\label{eq:tta_formula}
    \min_{\Tilde{\Theta}} \mL(\bx;\Theta),~ \bx \sim Q \left( \bx \right),
\end{equation}
where $\Tilde{\Theta} \subseteq \Theta$ denotes the free model parameters that should be updated. In general, the test-time learning objective $\mL(\cdot)$ can be formulated  as an entropy minimization problem~\cite{wang2021tent} or prediction consistency maximization over data augmentations~\cite{zhang2021memo}. 

For existing TTA methods like  TTT~\cite{sun2020test} and MEMO~\cite{zhang2021memo}), during the test-time adaptation, we shall need to compute one round or even multiple round of backward computation for each sample, which is very time-consuming and also not favorable for latency-sensitive applications. Moreover, most methods assume that all the test samples are drawn from OOD. However, in practice, the test samples may come from both ID and OOD. In fact, in many applications, the test set may contain a small portion of test samples. Simply optimizing the model on OOD test samples may lead to severe performance degradation in-distribution samples. We empirically validate the existence of such issue in Figure \ref{fig:imageC-forgetting-level-5}, where the updated model has a consistently lower accuracy on ID test samples than the original model.

\section{Proposed Methods}
In this paper, we propose an anti-forgetting test-time adaptation (\myeata) method, which aims  to improve the efficiency of test-time adaptation and tackle the catastrophic forgetting issue brought by existing TTA strategies  simultaneously.  
As shown in Figure~\ref{fig:overall_illustration}, \myeata consists of two strategies.  \textbf{1)}  \emph{Sample-efficient entropy minimization} (c.f. Section~\ref{sec:efficient_adaptation}) aims to conduct efficient adaptation relying on an active sample selection strategy. Here, the sample selection process is to choose only active samples for backward propagation and therefore  improve the overall TTA efficiency  (\ie, less gradient backward propagation). To this end, we devise an active sample selection score, denoted by $S(\bx)$, to detect those reliable and non-redundant test samples from the test set for TTA. \textbf{2)} \emph{Anti-forgetting weight regularization} (c.f. Section~\ref{sec:adaptation_wo_forgetting}) seeks to alleviate knowledge forgetting by enforcing that the parameters, important for the ID domain, do not change too much in test-time adaptation. In this way, the catastrophic forgetting issue can be significantly alleviated. The pseudo-code of \myeata is summarized in Algorithm~\ref{alg:overall}.

\subsection{Sample Efficient Entropy Minimization}\label{sec:efficient_adaptation}

For efficient test-time adaptation, we propose an active sample identification strategy to select  samples for backward propagation. Specifically, we design an active sample selection score for each sample, denoted by $S(\bx)$, based on two criteria: 1) samples should be \textbf{reliable} for test-time adaptation, and  2) the samples involved in optimization should be  \textbf{non-redundant}. By setting $S(\bx)\small{=}0$ for non-active samples, namely the unreliable and redundant samples, we can reduce unnecessary backward computation during test-time adaptation, thereby improving the prediction efficiency.

Relying on the sample score $S(\bx)$, following~\cite{wang2021tent,zhang2021memo}, we use entropy loss for model adaptations. Then, the sample-efficient entropy minimization is to minimize the following objective: 
\begin{align}
    \min_{\tilde{\Theta}}S(\bx) E(\bx;\Theta)\small{=-}S(\bx)\sum_{y\in\mC}f_{\Theta}(y|\bx)\text{log}f_{\Theta}(y|\bx), \label{eq:selective_entropy_optimization} 
\end{align}
where $\mC$ is the model output space. Here, the entropy loss $E(\cdot)$ is computed over a batch of samples each time (similar to Tent~\cite{wang2021tent}) to avoid a trivial solution, \ie, assigning all probability to the most probable class. For efficient adaptation, we update $\tilde{\Theta}\small{\subseteq}\Theta$ with the affine parameters of all batch normalization layers.

\begin{algorithm}[t]
	\caption{The pipeline of proposed \myeata.}
	\label{alg:overall}
	\begin{algorithmic}[1]
    \REQUIRE{Test samples $\mD_{test}\small{=}\{\bx_j\}_{j=1}^{M}$, the trained model $f_{\Theta}(\cdot)$, ID samples $\mD_{F}\small{=}\{\bx_q\}_{q=1}^Q$, batch size $B$.}
    \FOR{a batch $\mX\small{=}\{\bx_b\}_{b=1}^{B}$ in $\mD_{test}$}
    \STATE Calculate predictions $\hat{y}$ for all $\bx \in \mX$ via $f_{\Theta}(\cdot)$.
    \STATE Calculate sample selection score $S(\bx)$ via Eqn.~(\ref{eq:lambda_2nd}).
    \STATE Update model ($\tilde{\Theta} \small{\subseteq}\Theta$) with Eqn.~(\ref{eq:selective_entropy_optimization}) or Eqn.~(\ref{eq:overall_loss}).
    \ENDFOR
    \ENSURE The predictions $\{\hat{y}\}_{j=1}^M$ for all $\bx \in \mD_{test}$.
	\end{algorithmic}
\end{algorithm}

\begin{figure}[t]
\centering\hspace{-0.1in}
\includegraphics[width=0.9\linewidth]{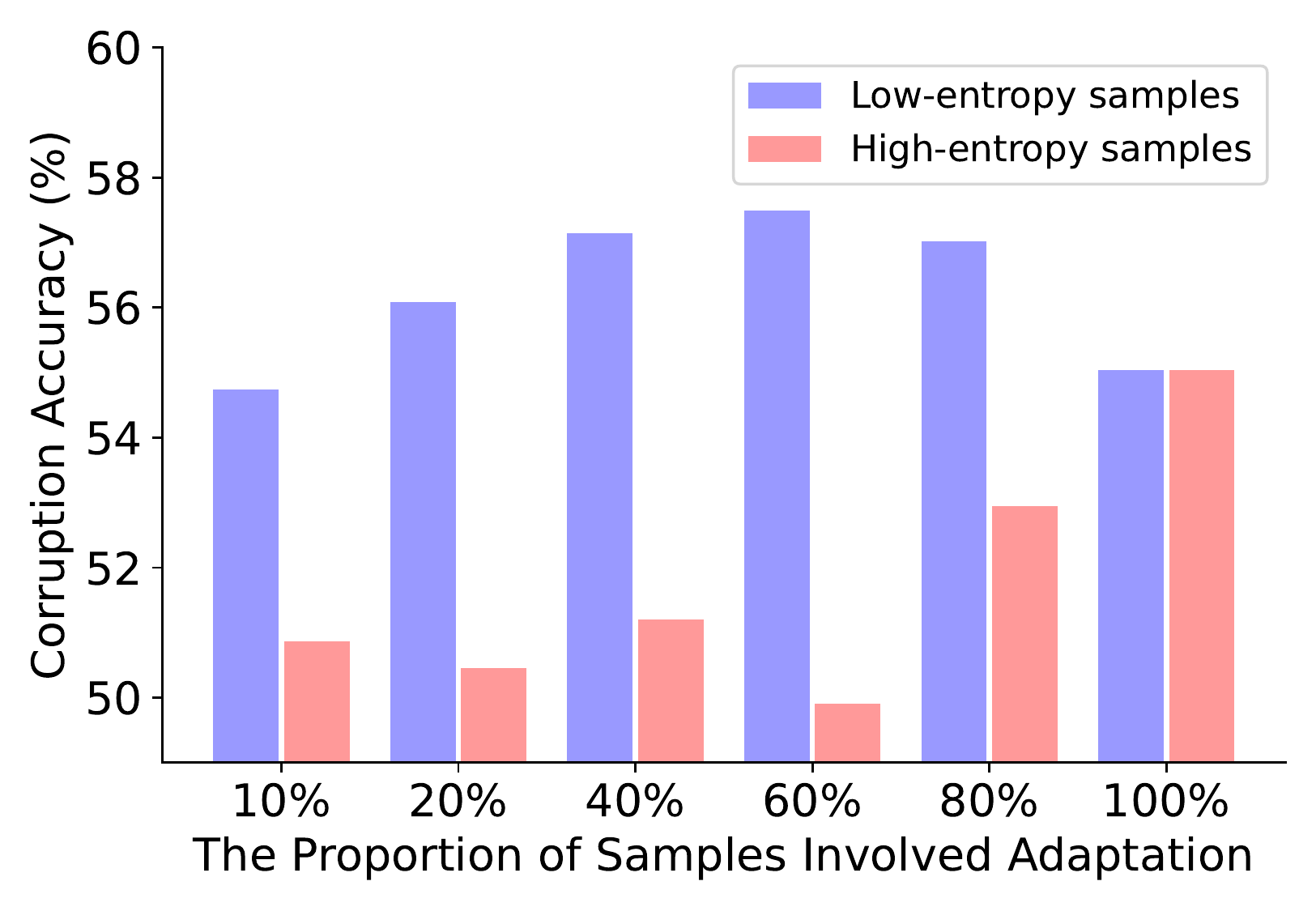}\vspace{-0.1in}
\caption{Effect of different test samples in test-time entropy minimization~\cite{wang2021tent}. We adapt a model on partial samples (top $p$\% samples with the highest or lowest entropy values), and then evaluate the adapted model on all test samples. Results are obtained on ImageNet-C (Gaussian noise, level 3) and ResNet-50 (base accuracy is 27.6\%). Introducing more samples with high entropy values into adaptation will hurt the adaptation performance.}
\label{fig:selective_entropy_motivation}
\end{figure}

\textbf{Reliable Sample Identification.}
Our intuition is that different test samples produce various effects in adaptation. To verify this, we conduct a preliminary study, where we select different proportions of samples (the samples are pre-sorted according to their entropy values $E(\bx;\Theta)$) for adaptation, and the resulting model is evaluated on all test samples. From Figure~\ref{fig:selective_entropy_motivation}, we find that: 1) adaptation on low-entropy samples makes more contribution than high-entropy ones, and 2) adaptation on test samples with very high entropy may hurt performance. The possible reason is that predictions of high-entropy samples are uncertain, so their gradients produced by entropy loss may be biased and unreliable. Following this, we name these low-entropy samples as reliable samples.

Based on the above observation, we propose an entropy-based weighting scheme to identify reliable samples and emphasize their contributions during adaptation. Formally, the entropy-based weight is given by:
\begin{align}
    S^{ent}(\bx) = \frac{1}{\exp \left[ E(\bx;\Theta) - E_0 \right]} \cdot \mathbb{I}_{\{E(\bx;\Theta)<E_0\}}(\bx), \label{eq:lambda_1st}
\end{align}
where $\mathbb{I}_{\{\cdot\}}(\cdot)$ is an indicator function,    $E(\bx;\Theta)$ is the entropy of sample $\bx$, and $E_0$ is a pre-defined threshold. The above weighting function excludes high-entropy samples from adaptation and assigns higher weights to test samples with lower prediction uncertainties, allowing them to contribute more to model updates. Note that evaluating $S^{ent}(\bx)$ does not involve any gradient back-propagation.

\textbf{Non-redundant Sample Identification.}
Although Eqn.~(\ref{eq:lambda_1st}) helps to exclude partial unreliable samples, the remaining test samples may still have redundancy. For example, given two test samples that are mutually similar and both have a lower prediction entropy than $E_0$, we still need to perform gradient back-propagation for each of them according to Eqn.~(\ref{eq:lambda_1st}). However, this is unnecessary as these two samples produce similar gradients for model adaptation.

To further improve efficiency, we propose to exploit the samples that can produce different gradients for model adaptation. Recall that the entropy loss only relies on final model outputs (\ie, classification logits),  we further filter samples by ensuring the remaining samples have diverse model outputs. To this end, one straightforward method is to save the model outputs of all previously seen samples, and then compute the similarity between the outputs of incoming test samples and all saved model outputs for filtering. However, this method is computational expensive at test time and memory-consuming with the increase of test samples. 

To address this, we exploit an exponential moving average technique to track the average model outputs of all seen test samples used for model adaptation. To be specific, given a set of model outputs of test samples, the moving average vector is updated recursively:
\begin{align}\label{eq:moving_average}
\bmm^t=\left\{\begin{array}{ll}
\bar{\by}^1, & \text { if } t=1 \\
\alpha\bar{\by}^t + (1-\alpha)\bmm^{t-1}, & \text { if } t>1
\end{array},\right.
\end{align}
where $\bar{\by}^t=\frac{1}{n}\sum_{k=1}^n\hat{\by}_k^t$ is the average model prediction of a mini-batch of $n$ test samples at the iteration $t$, and $\alpha\in[0,1]$. Following that, given a new test sample $\bx$ received at iteration $t > 1$, we compute the cosine similarity between its prediction $f_{\Theta}(\bx)$ and the moving average $\bmm^{t-1}$ (\ie, $cos(f_{\Theta}(\bx), \bmm^{t-1})$), which is then used to determine the diversity-based weight:
\begin{equation}
S^{div} \left( \bx \right) = \mathbb{I}_{\{cos(f_{\Theta}(\bx), \bmm^{t-1}) <\epsilon\}}(\bx),
\label{eq:lambda_div}
\end{equation}
where $\epsilon$ is a pre-defined threshold for cosine similarities. The overall sample-adaptive weight is then given by:
\begin{equation}\label{eq:lambda_2nd}
    S \left( \bx \right) = S^{ent} \left( \bx \right) \cdot S^{div} \left( \bx \right),
\end{equation}
which combines both entropy-based (as in Eqn.~\ref{eq:lambda_1st}) and diversity-based terms (as in Eqn.~\ref{eq:lambda_div}). Since we only perform gradient back-propagation for test samples with $S(\bx)>0$, the algorithm efficiency is further improved.

\textbf{Remark.} Given $M$ test samples $\mD_{test}=\{\bx_j\}_{j=1}^{M}$, the total number of reduced backward computation is given by $\mathbb{E}_{\bx\sim\mD_{test}} [\mathbb{I}_{\{S(\bx)=0\}}(\bx)]$, which is jointly determined the test data distribution $\mD_{test}$, entropy threshold $E_{0}$, and cosine similarity threshold $\epsilon$.

\subsection{Anti-Forgetting with Fisher Regularization}\label{sec:adaptation_wo_forgetting}

In this section, we propose a new weighted Fisher regularizer (called anti-forgetting regularizer) to alleviate the catastrophic forgetting issue caused by test-time adaptation, \ie, the performance of a test-time adapted model may significantly degrade on in-distribution (ID) test samples. We achieve this through weight regularization, which only affects the loss function and does not incur any additional computational overhead for model adaptation. To be specific, we apply an importance-aware regularizer $\mR$ to prevent model parameters, important for the in-distribution domain, from changing too much during the test-time adaptation process~\cite{kirkpatrick2017overcoming}:
\begin{equation}\label{eq:weight_regularization}
    \mR(\tilde{\Theta},\tilde{\Theta}^o)=\sum_{\theta_i\in\tilde{\Theta}}\omega(\theta_i)(\theta_i - \theta_i^{o})^2,
\end{equation}
where $\tilde{\Theta}$ are parameters used for model update and $\tilde{\Theta}^o$ are the corresponding parameters of the original model. $\omega(\theta_i)$ denotes the importance of $\theta_i$ and we measure it via the diagonal Fisher information matrix as in elastic weight consolidation~\cite{kirkpatrick2017overcoming}. Here, the calculation of Fisher information $\omega(\theta_i)$ is non-trivial since we are inaccessible to any labeled training data. For the convenience of presentation, we leave the details of calculating $\omega(\theta_i)$ in the next subsection.

After introducing the anti-forgetting regularizer, the final optimization formula for our method can be formulated as:
\begin{equation}\label{eq:overall_loss}
    \min_{\tilde{\Theta}} S(\bx) E(\bx;\Theta) + \beta \mR(\tilde{\Theta},\tilde{\Theta}^o),
\end{equation}
where $\beta$ is a trade-off parameter, $S(\bx)$ and $E(\bx;\Theta)$ are defined in Eqn.~(\ref{eq:selective_entropy_optimization}).

\textbf{Measurement of Weight Importance $\omega(\theta_i)$.} 
The calculation of Fisher information typically involves a set of labeled ID  training samples. However, in our problem setting, we are inaccessible to training data and the test samples are only unlabeled, which makes it non-trivial to measure the weight importance. To conquer this, we first collect a small set of unlabeled ID test samples $\{\bx_q\}_{q=1}^Q$, and then use the original trained model $f_{\Theta}(\cdot)$ to predict all these samples for obtaining the corresponding hard pseudo-label $\hat{y}_q$. Following that, we construct a pseudo-labeled ID test set $\mD_{F}=\{\bx_q, \hat{y}_q\}_{q=1}^Q$, based on which we calculate the fisher importance of model weights by:
\begin{equation}\label{eq:fisher_information}
    \omega(\theta_i) = \frac{1}{Q}\sum_{\bx_q\in\mD_{F}}\big(\frac{\partial}{\partial\theta_i^{o}}\mL_{CE}(f_{\Theta^o}(\bx_q),\hat{y}_q)\big)^2,
\end{equation}
where $\mL_{CE}$ is the cross-entropy loss. Here, we only need to calculate $\omega(\theta_i)$ once before performing test-time adaptation. 
\camera{Once calculated, we keep $\omega(\theta_i)$ fixed and apply it to any types of distribution shifts.}
Moreover, the unlabeled ID test samples are collected based on out-of-distribution detection techniques~\cite{liu2020energy,berger2021confidence}, which are easy to implement. Note that there is no need to collect too many ID test samples for calculating $\omega(\theta_i)$, \eg, 500 samples are enough for ImageNet-C dataset. More empirical studies regarding this can be found in Figure~\ref{fig:number_samples_fisher}.

\section{Experiments}

\begin{table*}[t]
    \caption{Comparison with state-of-the-art methods on ImageNet-C with the highest severity level 5 regarding \textbf{Corruption Error (\%, $\downarrow$)}.
    ``GN" and ``BN" denote group and batch normalization, respectively. ``JT" denotes the model is jointly trained via supervised cross-entropy and rotation prediction losses. The \textbf{bold} number indicates the best result and the \underline{underlined} number indicates the second best result.
    }
    \label{tab:imagenet-c-level-5}
\newcommand{\tabincell}[2]{\begin{tabular}{@{}#1@{}}#2\end{tabular}}
 \begin{center}
 \begin{threeparttable}
    \resizebox{1.0\linewidth}{!}{
 	\begin{tabular}{|l|ccc|cccc|cccc|cccc|cc|}
 	\multicolumn{1}{c}{} & \multicolumn{3}{c}{Noise} & \multicolumn{4}{c}{Blur} & \multicolumn{4}{c}{Weather} & \multicolumn{4}{c}{Digital} & \multicolumn{2}{c}{Average} \\
 	\toprule
 	 Method & Gauss. & Shot & Impul. & Defoc. & Glass & Motion & Zoom & Snow & Frost & Fog & Brit. & Contr. & Elastic & Pixel & JPEG & \#Forwards & \#Backwards \\
 	\midrule

        R-50 (GN)+JT   & 94.9 & 95.1 & 94.2 & 88.9 & 91.7 & 86.7 & 81.6 & 82.5 & 81.8 & 80.6 & 49.2 & 87.4 & 76.9 & 79.2 & 68.5 & 50,000 & 0\\
~~$\bullet~$TTT         & 69.0 & 66.4 & 66.6 & 71.9 & 92.2 & 66.8 & 63.2 & 59.1 & 81.0 & 49.0 & 38.2 & 61.1 & 50.6 & 48.3 & 52.0 & 50,000$\times$21 & 50,000$\times$20 \\
        R-50 (BN)   & 97.8 & 97.1 & 98.2 & 82.1 & 90.2 & 85.2 & 77.5 & 83.1 & 76.7 & 75.6 & 41.1 & 94.6 & 83.1 & 79.4 & 68.4 & 50,000 & 0\\
~~$\bullet~$TTA          & 95.9 & 95.1 & 95.5 & 87.5 & 91.8 & 87.1 & 74.2 & 86.0 & 80.9 & 78.7 & 47.0 & 87.6 & 85.4 & 75.4 & 66.4 & 50,000$\times$64 & 0 \\    
~~$\bullet~$BN adaptation          & 84.5 & 83.9 & 83.7 & 80.0 & 80.0 & 71.5 & 60.0 & 65.2 & 65.0 & 51.5 & 34.1 & 75.9 & 54.2 & 49.3 & 58.9 & 50,000 & 0 \\
~~$\bullet~$MEMO        & 92.5 & 91.3 & 91.0 & 80.3 & 87.0 & 79.3 & 72.4 & 74.7 & 71.2 & 67.9 & 39.0 & 89.0 & 76.2 & 67.0 & 62.5 & 50,000$\times$65 & 50,000$\times$64 \\
~~$\bullet~$Tent        & 71.6 & 69.8 & 69.9 & 71.8 & 72.7 & 58.6 & 50.5 & 52.9 & 58.7 & 42.5 & 32.6 & 74.9 & 45.2 & 41.5 & 47.7 & 50,000 & 50,000 \\
~~$\bullet~$Tent (episodic)  & 85.4 & 84.8 & 84.9 & 85.5 & 85.4 & 74.6 & 62.2 & 66.4 & 67.8 & 53.2 & 35.7 & 83.9 & 57.1 & 52.4 & 61.5 & 50,000$\times$2 & 50,000 \\
\midrule
~~$\bullet~$\myeta (ours) & \textbf{64.9} & \underline{62.1} & \underline{63.4} & \textbf{66.1} & 67.1 & \textbf{52.2} & 47.4 & \textbf{48.1} & \textbf{54.2} & \textbf{39.9} & 32.1 & \textbf{55.0} & \textbf{42.1} & \textbf{39.1} & \underline{45.1} & 50,000 & {26,031} \\ 
~~$\bullet~$\myeata (ours) & \underline{65.0} & 63.1 & 64.3 & 66.3 & \underline{66.6} & 52.9 & \underline{47.2} & \underline{48.6} & \underline{54.3} & \underline{40.1} & \textbf{32.0} & \underline{55.7} & \underline{42.4} & \underline{39.3} & \textbf{45.0} & 50,000 & 25,150 \\ 
~~$\bullet~$\myeata (lifelong) & \underline{65.0} & \textbf{61.9} & \textbf{63.2} & \underline{66.2} & \textbf{65.8} & \underline{52.7} & \textbf{46.8} & 48.9 & 54.4 & 40.3 & \textbf{32.0} & 55.8 & 42.8 & 39.6 & 45.3 & 50,000 & {28,243} \\ 
    \bottomrule
	\end{tabular}
	}
	 \end{threeparttable}
	 \end{center}
\end{table*}

We organize the experiments to answer following questions: 1) How does \myeata compare with prior methods regarding efficiency and accuracy? 2) Can \myeata alleviate the forgetting 
that occurred after test-time adaptation? 3) How do different components affect the performance of \myeata? 

\textbf{Datasets and Models.} 
We conduct experiments on three benchmarks datasets for OOD generalization, \ie, CIFAR-10-C, ImageNet-C~\cite{hendrycks2019benchmarking} (contains corrupted images in 15 types of 4 main categories and each type has 5 severity levels) and ImageNet-R~\cite{hendrycks2021many}.
We use ResNet-26 (R-26)/ResNet-50 (R-50)~\cite{he2016deep} for CIFAR-10/ImageNet experiments. The models are trained on CIFAR-10 or ImageNet training set and then tested on clean or the above OOD test sets.

\textbf{Compared Methods.}
We compare with following state-of-the-art methods. Test-time Training (TTT)~\cite{sun2020test} adapts a model via rotation prediction at test time, but requires the model also being trained by rotation prediction. Test-time Augmentation (TTA)~\cite{ashukha2020pitfalls} predicts a sample via the average outputs of its different augmentations. BN adaptation~\cite{schneider2020improving} updates batch normalization statistics on test samples. Tent~\cite{wang2021tent} minimizes the entropy of test samples during testing. MEMO~\cite{zhang2021memo} maximizes the prediction consistency of different augmented copies regarding a given test sample. We denote \myeata without weight regularization in Eqn.~(\ref{eq:weight_regularization}) as \textbf{e}fficient \textbf{t}est-time \textbf{a}daptation (\myeta).

\begin{figure*}[t]
\centering     
\subfigure{\label{fig:imageC-forgetting-level-5-each-reset}\includegraphics[width=83mm]{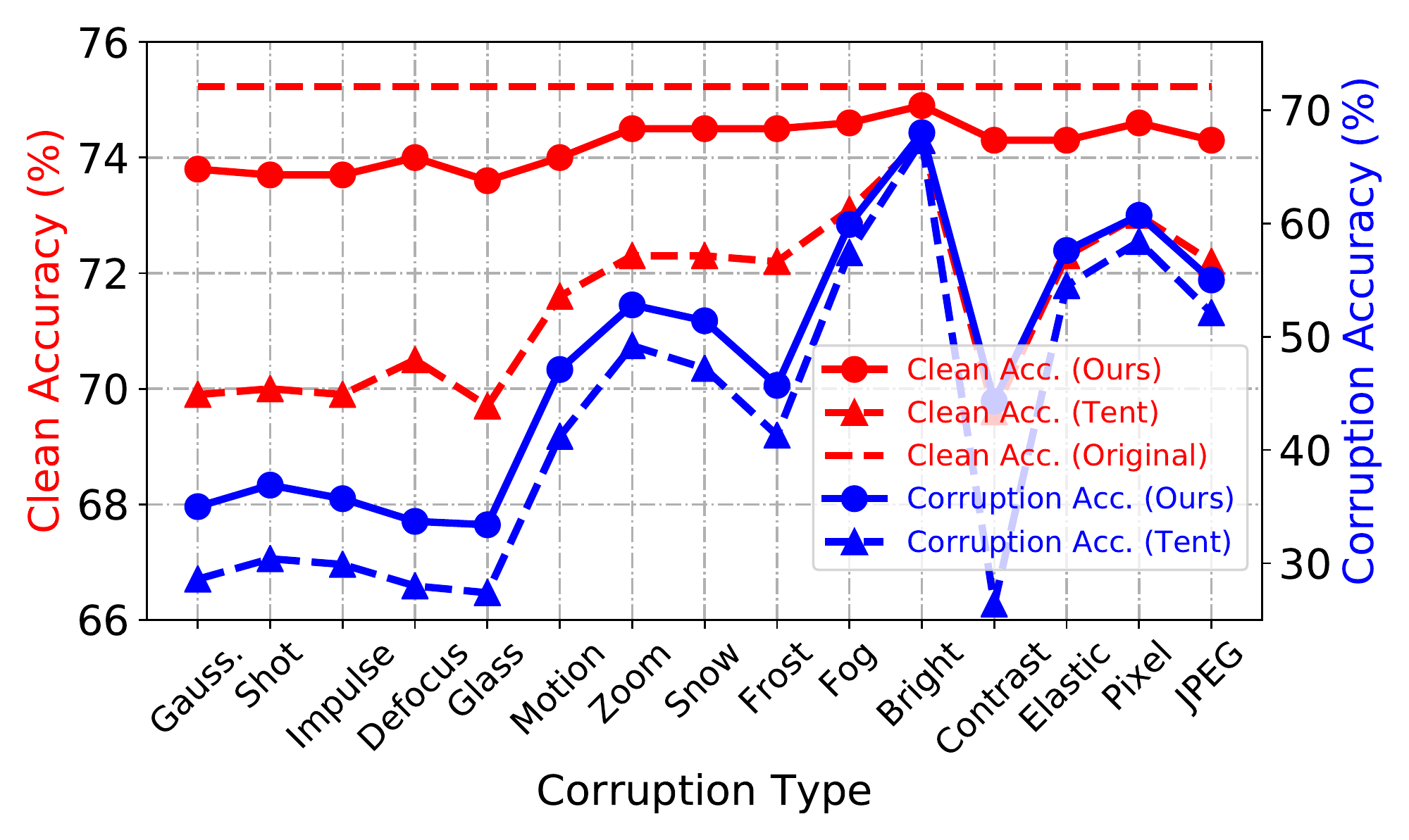}}
\subfigure{\label{fig:imageC-forgetting-level-5-lifelong}\includegraphics[width=82mm]{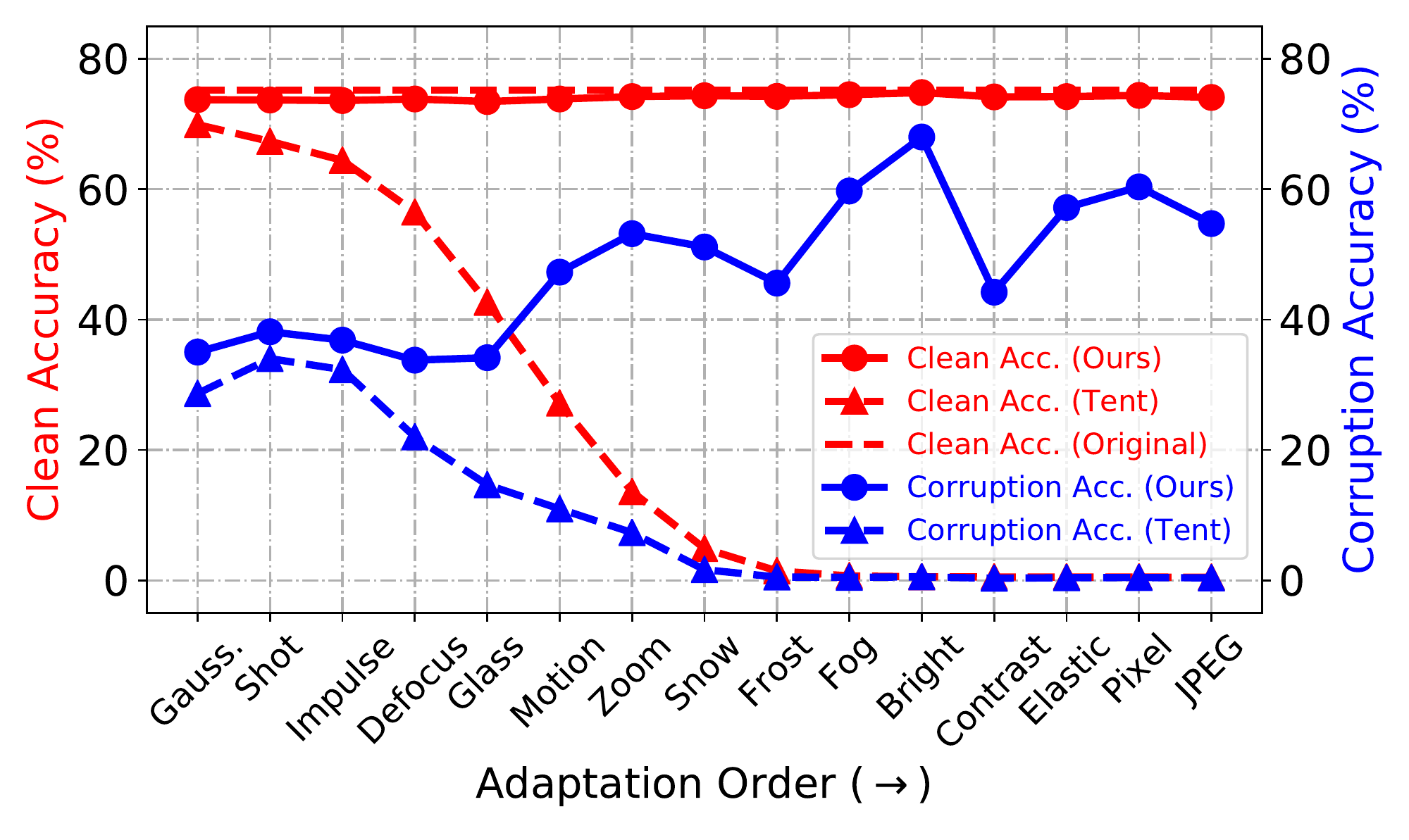}}
\vspace{-0.1in}
\caption{Comparison of prevent forgetting on ImageNet-C (severity level 5) with ResNet-50. We record the OOD corruption accuracy on each corrupted test set and the ID clean accuracy (after OOD adaptation).  In \textbf{Left}, the model parameters of both Tent and our \myeata are reset before adapting to a new corruption type. In \textbf{Right}, the model performs lifelong adaptation and the parameters will never be reset, namely Tent (lifelong) and our \myeata (lifelong). \myeata achieves higher OOD accuracy and meanwhile maintains the ID clean accuracy. }
\label{fig:imageC-forgetting-level-5}
\end{figure*}

For TTT, Tent, our \myeta and \myeata, the model is online adapted through the entire evaluation of one given test dataset (\eg, gaussian noise level 5 of ImageNet-C). Once the adaptation on this dataset is finished, the model parameters will be reset. For TTT (episodic) and Tent (episodic), the model parameters will be reset immediately after each optimization step of a test sample or batch. For \myeata (lifelong) and Tent (lifelong), the model is online adapted and the parameters will never be reset (as shown in Figure~\ref{fig:imageC-forgetting-level-5} (\textbf{Right})), which is more challenging but practical.

\textbf{Evaluation Metrics.}
1) Clean accuracy/error (\%) on original in-distribution (ID) test samples, \eg, the original testing images of ImageNet. \camera{Note that we measure the clean accuracy of all methods via (re)adaptation}; 2) Out-of-distribution (OOD) accuracy/error (\%) on OOD test samples, \eg, the corruption accuracy on ImageNet-C; 3) The number of forward and backward passes during the entire test-time adaptation process.
Note that the fewer \#forwards and \#backwards indicate the less computation, leading to higher efficiency.

\textbf{Implementation Details.} 
For test-time adaptation, we use SGD as the update rule, with a momentum of 0.9, batch size of 64, and learning rate of 0.005/0.00025 for CIFAR-10/ImageNet (following Tent and MEMO). The entropy constant $E_0$ in Eqn.~(\ref{eq:lambda_1st}) is set to $0.4\times\ln C$, where $C$ is number of task classes. The $\epsilon$ in Eqn.~(\ref{eq:lambda_2nd}) is set to 0.4/0.05 for CIFAR-10/ImageNet. The trade-off parameter $\beta$ in Eqn.~(\ref{eq:overall_loss}) is set to 1/2,000 for CIFAR-10/ImageNet to make two losses have the similar magnitude. We use 2,000 samples to calculating $\omega(\theta_i)$ in Eqn.~(\ref{eq:fisher_information}) \camera{which takes 2,000 additional forward-and-backward passes.} The moving average factor $\alpha$ in Eqn.~(\ref{eq:moving_average}) is set to 0.1.
More details are put in Supplementary.

\subsection{Comparisons of OOD Performance and Efficiency}

\begin{table}[t]
    \caption{Comparison on ImageNet-R. The base model is ResNet-50 (using batchnorm) trained on original ImageNet training set.}
	\label{tab:imagenet-r}
\newcommand{\tabincell}[2]{\begin{tabular}{@{}#1@{}}#2\end{tabular}}
 \begin{center}
 \begin{threeparttable}
    \resizebox{1.0\linewidth}{!}{
 	\begin{tabular}{|l|ccc|}\toprule
        \multirow{1}{*}{\tabincell{c}{Model}}&\multirow{1}{*}{\tabincell{c}{Error (\%)}}&\multirow{1}{*}{\tabincell{c}{\#Forwards}}&\multirow{1}{*}{\tabincell{c}{\#Backwards}} \\
        
        \midrule
         Base Model &63.8	&30,000 &0      \\
         ~~$\bullet~$TTA~\cite{ashukha2020pitfalls}  &61.3$_{(-2.5)}$	     & 30,000$\times$64  & 0     \\
         ~~$\bullet~$BN~\cite{schneider2020improving}   &59.7$_{(-4.1)}$        & 30,000            & 0      \\
         ~~$\bullet~$MEMO~\cite{zhang2021memo} &58.8$_{(-5.0)}$	     & 30,000$\times$65  & 30,000$\times$64      \\
         ~~$\bullet~$Tent~\cite{wang2021tent} &57.7$_{(-6.1)}$	     & 30,000            & 30,000      \\
         ~~$\bullet~$Tent (episodic) &61.0$_{(-2.9)}$	     & 30,000            & 30,000      \\
        \midrule
          ~~$\bullet~$\myeta (ours) &\textbf{54.5$_{(-\textbf{9.3})}$}	 & 30,000            & 14,847      \\ 
          ~~$\bullet~$\myeata (ours) &54.8$_{(-9.0)}$	 & 30,000            & 14,800      \\ 
        \bottomrule
	\end{tabular}}
	 \end{threeparttable}
	 \end{center}
	 \vspace{-0.1in}
\end{table}

\textbf{Results on ImageNet-C.}
We report the comparisons on ImageNet-C with the highest severity level 5 in Table~\ref{tab:imagenet-c-level-5} and put more results of other severity levels 1-4 into Supplementary due to the page limitation. From the results, our \myeta and \myeata consistently outperform the considered methods in all 15 corruption types regarding the classification error, suggesting our effectiveness. With our sample-adaptive entropy loss, \myeta achieves a large performance gain over Tent (\eg, $71.6\%\rightarrow65.0\%$ on Gaussian noise), verifying that removing samples with unreliable gradients and tackling samples differently benefits the test-time adaptation. More critically, \myeta outperforms TTT consistently (while Tent fails to achieve this), demonstrating the potential of fully test-time adaptation methods, \ie, boosting OOD generalization without altering the training process. 
Compared with \myeta, \myeata and \myeata (lifelong) achieve comparable OOD performance (but prevent the forgetting on ID samples, see Figure~\ref{fig:imageC-forgetting-level-5}), showing that our anti-forgetting regularization does not limit the learning ability during adaptation.

As for efficiency, the required average backward number of our \myeta is 26,031, which is much fewer than those methods that need multiple data augmentations (\ie, TTT and MEMO are 50,000$\times$20 and 64). Compared with Tent, \myeta reduces the average \#backward from 50,000 to 26,031, by excluding samples with high prediction entropy and samples that are similar out of test-time optimization. In this sense, our method only needs to adapt for partial samples, resulting in higher efficiency. Note that although optimization-free methods (such as BN adaptation) do not need backward updates, their OOD performances are limited.

\begin{table}[t]
	\caption{
	Comparison on CIFAR-10-C. Each result is averaged over 15 different corruption types and 5 severity levels (totally 75).}
	\label{tab:cifar-10-c}
\newcommand{\tabincell}[2]{\begin{tabular}{@{}#1@{}}#2\end{tabular}}
 \begin{center}
 \begin{threeparttable}
    \resizebox{1.0\linewidth}{!}{
 	\begin{tabular}{|l|ccc|}\toprule
        \multirow{2}{*}{\tabincell{c}{Model}}&\multirow{2}{*}{\tabincell{c}{Average \\ Error (\%)}}&\multirow{2}{*}{\tabincell{c}{Average \\ \#Forwards}}&\multirow{2}{*}{\tabincell{c}{Average \\ \#Backwards}} \\
        ~ & ~ & ~ & ~ 
        \\
        \midrule
         ResNet-26 (GroupNorm) &22.5	&10,000 &0      \\
         ~~$\bullet~$TTA~\cite{ashukha2020pitfalls} &19.9$_{(-2.6)}$	&10,000$\times$32 &0     \\
         ~~$\bullet~$MEMO~\cite{zhang2021memo} &19.6$_{(-2.9)}$	&10,000$\times$33 &10,000$\times$32      \\
         ResNet-26 (GroupNorm)+JT &22.8	&10,000  &0   \\
         ~~$\bullet~$TTT~\cite{sun2020test} & \textbf{15.6}$_{(-7.2)}$	&10,000$\times$33 &10,000$\times$32      \\
         ~~$\bullet~$TTT (episodic) & 21.5$_{(-1.3)}$	&10,000$\times$33 &10,000$\times$32      \\
         ResNet-26 (BatchNorm) &28.4	&10,000    &0  \\
         ~~$\bullet~$Tent~\cite{wang2021tent} &20.2$_{(-8.2)}$ &10,000	&10,000      \\
         \midrule
         ~~$\bullet~$\myeta (ours) &19.4$_{(-\textbf{9.0})}$ &10,000	&8,192      \\
         ~~$\bullet~$\myeata (ours) &19.7$_{(-8.7)}$ &10,000	&8,153      \\
        \bottomrule
	\end{tabular}}
	 \end{threeparttable}
	 \end{center}
\end{table}

\textbf{Results on ImageNet-R and CIFAR-10-C.} From Table~\ref{tab:imagenet-r}, our \myeta yields 54.5\% classification error on ImageNet-R (-3.2\% over the best counterpart method Tent) and only needs 14,847 backward propagation (much fewer than other learning-based test-time adaptation methods, \eg, MEMO and Tent). The results on CIFAR-10-C are shown in Table~\ref{tab:cifar-10-c}. Under the same base model (ResNet-26 with batch normalization), \myeta achieves lower average error than Tent (19.4\% \vs~20.2\%) with less requirements of back-propagation (8,192 \vs~10,000). Moreover, the performance gain over the base model of \myeta is larger than that of TTT, \ie, -9.0\% \vs~-7.2\% average error. These results are consistent with the ones on ImageNet-C that \myeta achieves higher performance and improves the efficiency, further demonstrating the effectiveness and superiority of our method.

\begin{figure}[t]
\centering     
\subfigure[Effect of different entropy margins $E_0$ in Eqn.~(\ref{eq:lambda_1st})]{\label{fig:entropy0_effects}\includegraphics[width=75mm]{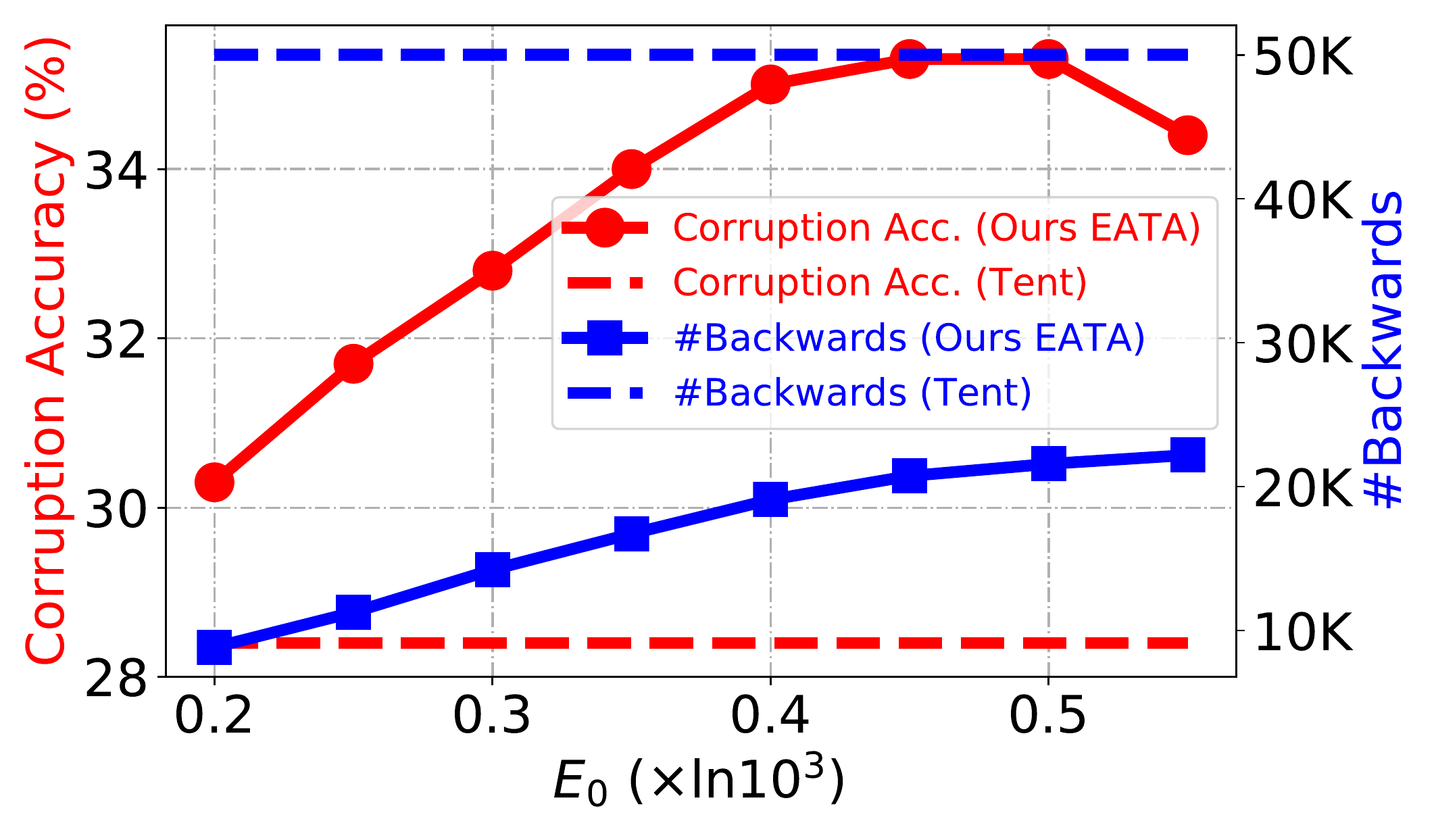}}
\hspace{0.2in}
\subfigure[Effect of \#samples for calculating Fisher in Eqn.~(\ref{eq:fisher_information}).]{\label{fig:number_samples_fisher}\includegraphics[width=75mm]{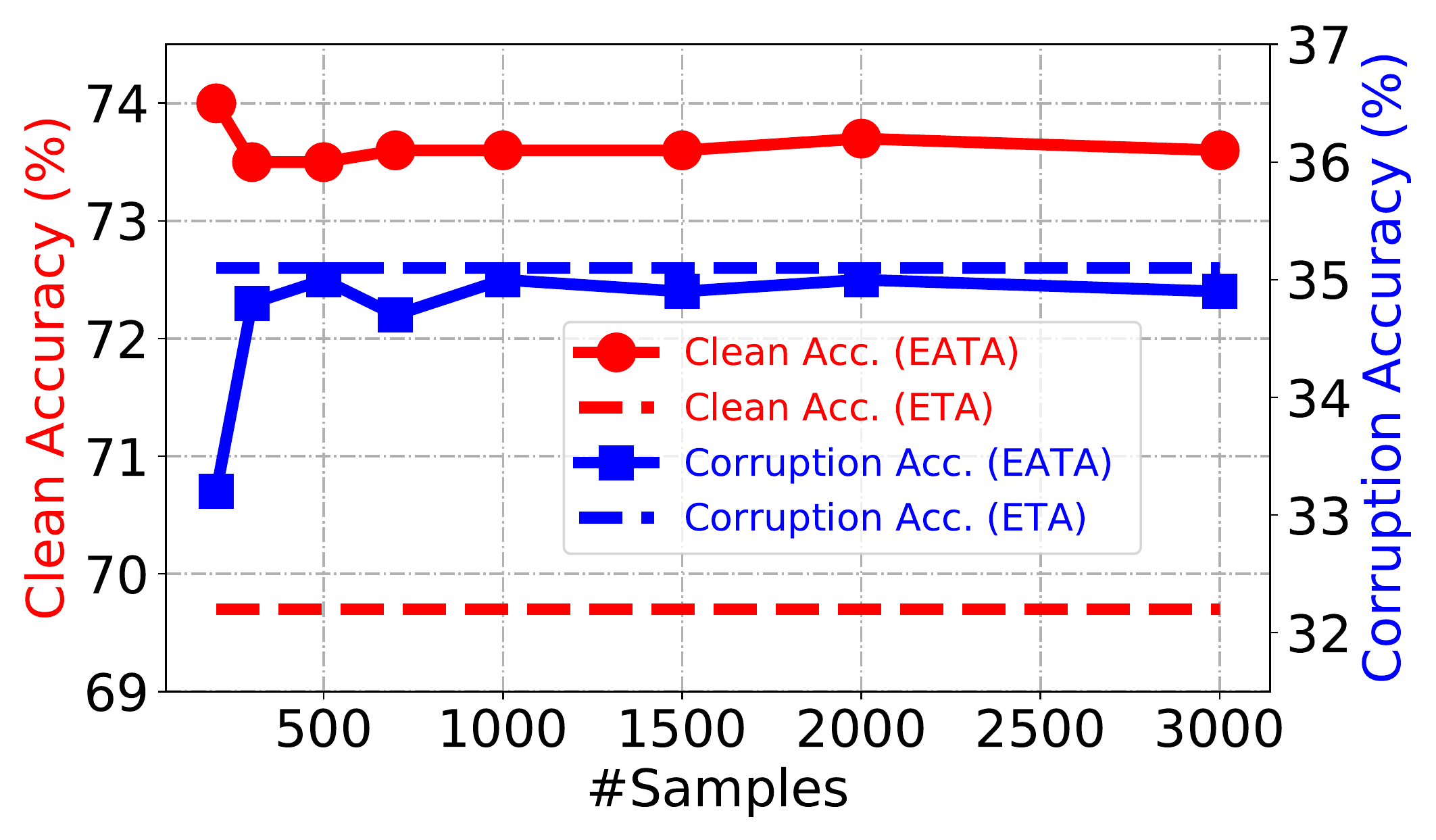}}
\vspace{-0.1in}
\caption{Ablation experiments on ImageNet-C (Gaussian noise, severity level=5) with ResNet-50.}
\label{fig:ablation_e0_fisher_num}
\end{figure}

\subsection{Demonstration of Preventing Forgetting}

In this section, we investigate the ability of our \myeata in preventing ID forgetting during test-time adaptation. The experiments are conducted on ImageNet-C with ResNet-50. We measure the anti-forgetting ability by comparing the model's clean accuracy (\ie, on original validation data of ImageNet) before and after adaptation. To this end, we first perform test-time adaptation on a given OOD test set, and then evaluate the clean accuracy of the updated model. Here, we consider two adaptation scenarios: i) the model parameters will be reset before adapting to a new corrupted test set; 2) the model parameters will never be reset (namely lifelong adaptation), which is more challenging but practical. We report the results of severity level 5 in Figure~\ref{fig:imageC-forgetting-level-5} and put results of severity levels 1-4 into Supplementary.

From Figure~\ref{fig:imageC-forgetting-level-5}, our \myeata consistently outperforms Tent regarding the OOD corruption accuracy and meanwhile maintains the clean accuracy on ID data (in both two adaptation scenarios), demonstrating our effectiveness. It is worth noting that the performance degradation in lifelong adaptation scenario is much more severe (see Figure~\ref{fig:imageC-forgetting-level-5} \textbf{Right}). 
More critically, in lifelong adaptation, both the clean and corruption accuracy of Tent decreases rapidly (until degrades to 0\%) after adaptation of the first three corruption types, showing that Tent in lifelong adaptation is not stable enough. In contrast, during the whole lifelong adaptation process, our \myeata achieves good corruption accuracy and the clean accuracy is also very close to the clean accuracy of the model without any OOD adaptation (\ie, original clean accuracy, tested using Tent). These results demonstrate the superiority of \myeata in terms of overcoming the forgetting on ID data.

\subsection{Ablation Studies}

\textbf{Effect of Components in $S(\bx)$ (Eqn.~\ref{eq:lambda_2nd}).}
Our \myeata accelerates test-time adaptation by excluding two types of samples out of optimization: 1) samples with high prediction entropy values (Eqn.~\ref{eq:lambda_1st}) and 2) samples that are similar (Eqn.~\ref{eq:lambda_2nd}). We ablate both of them in Table~\ref{tab:effects_diff_components}. Compared with the baseline $S(\bx)\small{=}1$ (the same as Tent), introducing $S^{ent}(\bx)$ in Eqn.~(\ref{eq:lambda_1st}) achieves lower error and fewer backwards (\eg, 65.7\% (26,694) \vs~71.6\% (50,000) on level 5). This verifies our motivation in Figure~\ref{fig:selective_entropy_motivation} that some high-entropy samples may hurt the performance since their gradients are unreliable.
When further removing some redundant samples that are similar (Eqn.~\ref{eq:lambda_2nd}), our \myeata further reduces the number of back-propagation (\eg, 26,694$\rightarrow$19,121 on level 5) and achieves comparable OOD error (\eg, 65.0\% \vs~65.7\%), demonstrating the effectiveness of our proposed sample-efficient optimization strategy. 

\textbf{Entropy Constant $E_0$ in Eqn.~(\ref{eq:lambda_1st}).}
We evaluate our \myeata with different $E_0$, selected from \{0.20, 0.25, 0.30, 0.35, 0.40, 0.45, 0.50, 0.55\}$\times\ln 10^3$, where $10^3$ is the class number of ImageNet. From Figure~\ref{fig:entropy0_effects}, our \myeata achieves excellent performance when $E_0$ belongs to $[0.4, 0.5]$. Either a smaller or larger $E_0$ would hamper the performance. The reasons are mainly as follows. When $E_0$ is small, \myeata removes too many samples during adaptation and thus is unable to learn enough adaptation knowledge from the remaining samples. When $E_0$ is too large, some high-entropy samples would take part in the adaptation but contribute unreliable and harmful gradients, resulting in performance degradation. As for larger $E_0$ leads to more backward passes, we set $E_0$ to 0.4$\times\ln 10^3$ for the efficiency-performance trade-off and fix the proportion of 0.4 for all other experiments.

\begin{table}[t]

\vspace{-0.1in}
	\caption{
	Effectiveness of components in sample-adaptive weight $S(\bx)$ in \myeata on ImageNet-C (Gaussian noise) with ResNet-50.
	}
	\label{tab:effects_diff_components}
\newcommand{\tabincell}[2]{\begin{tabular}{@{}#1@{}}#2\end{tabular}}
 \begin{center}
 \begin{threeparttable}
    \resizebox{1.0\linewidth}{!}{
 	\begin{tabular}{|l|cccc|}\toprule
        \multicolumn{1}{|c|}{\multirow{2}[0]{*}{Method}} & \multicolumn{2}{c}{Level 3} & \multicolumn{2}{c|}{Level 5} \\
         & \multicolumn{1}{c}{Error (\%)} & \multicolumn{1}{c}{\#Backwards} & \multicolumn{1}{c}{Error (\%)} & \multicolumn{1}{c|}{\#Backwards} \\
        \midrule
         Baseline ($S(\bx)\small{=}1$) & 45.3  & 50,000	& 71.6  & 50,000      \\
         ~~$+S^{ent}{(\bx)}$ (Eqn.~\ref{eq:lambda_1st}) &43.0 & 37,943	     & 65.7 & 26,694        \\
         ~~$+S{(\bx)}$ (Eqn.~\ref{eq:lambda_2nd})  &\textbf{42.6} &\textbf{29,051}       & \textbf{65.0} & \textbf{19,121}                 \\
        \bottomrule
	\end{tabular}}
	 \end{threeparttable}
	 \end{center}
	
\vspace{-0.1in}
\end{table}

\textbf{Number of Samples for Calculating Fisher in Eqn.~(\ref{eq:fisher_information}).}
As described in Section~\ref{sec:adaptation_wo_forgetting}, the calculation of Fisher information involves a small set of unlabeled ID samples, which can be collected via existing OOD detection techniques~\cite{berger2021confidence}. Here, we investigate the effect of \#samples $Q$, selected from \{200, 300, 500, 700, 1000, 1500, 2000, 3000\}. From Figure~\ref{fig:number_samples_fisher}, our \myeata achieves stable performance with $Q\geq 300$, \ie, compared with \myeta (without regularization), the OOD performance is comparable and the clean accuracy is much higher. These results show that our \myeata does not need to collect too many ID samples, which are easy to obtain in practice.

\subsection{\camera{More Discussions}}

\textbf{Results on Single Sample Adaptation (Batch Size $B=1$).} In our proposed methods, we conduct test-time learning over a batch of test samples each time. When batch size $B$ equals to 1, directly applying Tent and our \myeta may fail (see Table~\ref{tab:bs=1}), but it is not a very significant issue in practice. Actually, one can address this by either maintaining a sliding window $\{\bx_i\}_{i=t-L}^{t}$ including $L$ previous samples and the current test sample $\bx_t$, or data augmentation techniques adopted in existing single sample TTA methods (\cite{zhang2021memo,khurana2021sita}). The first way is simple to implement and incurs only minor extra costs. From Table~\ref{tab:bs=1}, our \myeta with sliding window (\myeta-wnd) works well and consistently outperforms the Tent counterpart.

\begin{table}[ht]
    \caption{Effectiveness of \myeta under sliding window strategy (with different window length $L$) for single sample adaptation. We report corruption accuracy (\%) on ImageNet-C (Gaussian noise, level 5).}
	\label{tab:bs=1}
\newcommand{\tabincell}[2]{\begin{tabular}{@{}#1@{}}#2\end{tabular}}
 \begin{center}
 \begin{threeparttable}
    \resizebox{1.0\linewidth}{!}{
 	\begin{tabular}{|c|c|c|c|c|c|c|}
    \toprule
        \multirow{2}{*}{\tabincell{c}{Base ResNet-50}}&\multirow{2}{*}{\tabincell{c}{Tent}}&\multirow{2}{*}{\tabincell{c}{ETA}}&\multirow{2}{*}{\tabincell{c}{Tent-wnd \\ ($L\small{=}32)$ }}&\multirow{2}{*}{\tabincell{c}{ETA-wnd \\ ($L\small{=}32$)}}&\multirow{2}{*}{\tabincell{c}{Tent-wnd \\ ($L\small{=}64$) }}&\multirow{2}{*}{\tabincell{c}{ETA-wnd \\ ($L\small{=}64$)}} \\
        ~ &~ &~ &~ &~ & ~ & ~\\
        \midrule
         2.2 &0.1	&0.1 &28.1 &30.8 & 29.5 & \textbf{32.4}     \\
        \bottomrule
	\end{tabular}}
	 \end{threeparttable}
	 \end{center}
\end{table}

\textbf{Wall-clock Time Speed-up of \myeata}.
In the current PyTorch version, gradient computation is conducted on the full mini-batch, even if instance-wise masks are applied. We achieve wall-time speed-up with one forward-only pass with $B$ samples, and one forward-and-backward pass with $B'$ samples selected by \myeta/\myeata. For ResNet-50 on ImageNet-C (Gaussian noise, level=5, 50,000 images) with one Tesla V100 GPU, the actual run time is 113s for Tent (28.6\% accuracy) and 102s for \myeata (35.1\% accuracy). Actually, this is a more engineering problem, and an ideal implementation (forward with $B$ samples and backward with $B'$ samples) should further speed up the computation.

\textbf{Additional Memory by Fisher Regularizer.} 
Since we only regularize the affine parameters of BN layers, \myeata needs very little extra memory. For ResNet-50 on ImageNet-C, the extra GPU memory at run time is only 9.8 MB, which is much less than that of Tent with batch size 64 (5,675 MB).

\textbf{\myeata under Mixed-and-Shifted Distributions.} We evaluate Tent and our \myeata on mixed ImageNet-C (level=3 or 5) that consists of 15 different corruption types/distribution shifts (totally 750k images). Results in Table~\ref{tab:mixed_distribution} show the stability of \myeata on large-scale and complex TTA scenarios.   

\begin{table}[t!]
    \caption{Comparison with Tent~\cite{wang2021tent} \wrt corruption accuracy (\%) with mixture of 15 corruption types on ImageNet-C.}
	\label{tab:mixed_distribution}
\newcommand{\tabincell}[2]{\begin{tabular}{@{}#1@{}}#2\end{tabular}}
 \begin{center}
 \begin{threeparttable}
    \resizebox{1.0\linewidth}{!}{
 	\begin{tabular}{|l|c|c|c|}
 	\toprule
        \multirow{1}{*}{\tabincell{c}{~Severity~~}}&\multirow{1}{*}{\tabincell{c}{~~Base (ResNet-50)~~}}&\multirow{1}{*}{\tabincell{c}{~~~~~~~Tent~~~~~~~}}&\multirow{1}{*}{\tabincell{c}{~~\myeata (ours)~}}\\
        \midrule
         Level=3 &39.8	&41.5$_{(+1.7)}$ &\textbf{52.6}$_{(+12.8)}$      \\
         Level=5 &18.0	&2.3$_{(-15.7)}$ &\textbf{26.6}$_{(+8.6)}$      \\
        \bottomrule
	\end{tabular}}
	 \end{threeparttable}
	 \end{center}
\end{table}

\textbf{\myeata with Large Network Models.} In Table \ref{tab:big_models}, we apply \myeata to pre-trained models with different computational complexities.  From the results, even for large models (ResNet-101 and ResNet-152) that tend to show over-confident, our \myeata still works well.

\begin{table}[t!]
    \caption{Effectiveness of \myeata with different  backbone models. We report the corruption accuracy (\%) on ImageNet-C (Gaussian noise, severity level 5).} 
	\label{tab:big_models}
\newcommand{\tabincell}[2]{\begin{tabular}{@{}#1@{}}#2\end{tabular}}
 \begin{center}
 \begin{threeparttable}
    \resizebox{1.0\linewidth}{!}{
 	\begin{tabular}{|l|c|c|c|}
 	\toprule
        \multirow{1}{*}{\tabincell{c}{Base}}&\multirow{1}{*}{\tabincell{c}{~~ResNet-50~~ (2.2)}}&\multirow{1}{*}{\tabincell{c}{~~ResNet-101 (3.5)~~}}&\multirow{1}{*}{\tabincell{c}{~~ResNet-152 (3.6)~~}}\\
    \midrule
         Tent &28.4$_{(+26.2)}$	&32.5$_{(+29.0)}$ &33.7$_{(+30.1)}$      \\
         \myeata &\textbf{35.0$_{(+32.8)}$}	&\textbf{38.8$_{(+35.3)}$} &\textbf{40.2$_{(+36.6)}$}      \\
    \bottomrule
	\end{tabular}}
	 \end{threeparttable}
	 \end{center}
\end{table}

\textbf{\myeata with Different Number of Test Samples.} 
We investigate the effect of the total number of test samples ($N$) in \myeata, where the fewer samples are sampled from the entire test set. 
From Table~\ref{tab:small_testset}, \myeata works well and consistently outperforms Tent counterpart, regardless of the number of test samples. Meanwhile, if there are many test samples, \myeata would benefit more, \ie, larger accuracy gain.

\begin{table}[t!]
    \caption{Comparison with Tent~\cite{wang2021tent} \wrt corruption accuracy (\%) with fewer ($N$) test samples on ImageNet-C (Gaussian noise, severity level 5).}
	\label{tab:small_testset}
\newcommand{\tabincell}[2]{\begin{tabular}{@{}#1@{}}#2\end{tabular}}
 \begin{center}
 \begin{threeparttable}
    \resizebox{1.0\linewidth}{!}{
 	\begin{tabular}{|l|c|c|c|c|c|c|}
    \toprule
        \multirow{1}{*}{\tabincell{c}{Method}}&\multirow{1}{*}{\tabincell{c}{$N\small{=}$256}}&\multirow{1}{*}{\tabincell{c}{$N\small{=}$512}}&\multirow{1}{*}{\tabincell{c}{$N\small{=}$1,024}}&\multirow{1}{*}{\tabincell{c}{$N\small{=}$2,048}}&\multirow{1}{*}{\tabincell{c}{$N\small{=}$4,096}}&\multirow{1}{*}{\tabincell{c}{$N\small{=}$10,000}}\\
    \midrule
         Tent &13.7 &	18.2 &	14.8 &	15.2 &	17.5 &	20.9     \\
         \myeata &\textbf{14.8} &	\textbf{20.1} &	\textbf{16.3} &	\textbf{18.7} &	\textbf{23.5} & 	\textbf{27.6}     \\
    \bottomrule
	\end{tabular}}
	 \end{threeparttable}
	 \end{center}
\end{table}

\section{Conclusion}

In this paper, we propose an efficient anti-forgetting test-time adaptation method, to improve the performance of pre-trained models on a potentially shifted test domain. Specifically, we devise a sample-efficient entropy minimization strategy that selectively performs test-time optimization with reliable and non-redundant samples. This improves the adaptation efficiency and meanwhile boosts the out-of-distribution performance. In addition, we introduce a Fisher-based anti-forgetting regularizer into test-time adaptation. With this loss, a model can be adapted continually without performance degradation on in-distribution test samples, making test-time adaptation more practical for real-world applications. Extensive experimental results on several benchmark datasets demonstrate the effectiveness of our proposed method.

\section{Acknowledgments}
\camera{This work was partially supported by the 
Ministry of Science and Technology Foundation Project 2020AAA0106900, 
Key Realm R\&D Program of Guangzhou 202007030007,
National Natural Science Foundation of China (NSFC) 62072190, Program for Guangdong Introducing Innovative and Enterpreneurial Teams 2017ZT07X183.}

\balance
\bibliography{main}
\bibliographystyle{icml2022}

\newpage
\appendix
\onecolumn

\icmltitle{Supplementary Materials for \\``\mytitle''}

In the supplementary, we provide more implementation details and more experimental results of our \myeata.
We organize our supplementary as follows.

\begin{itemize}[leftmargin=*]
    \item In Section~\ref{supp:sec:more_impl}, we provide more experimental details of our proposed \myeata.
    \item In Section~\ref{supp:sec:results_perf_efc}, we show more experimental results to compare the out-of-distribution performance and efficiency with state-of-the-art methods on ImageNet-C with different corruption types and severity levels.
    \item In Section~\ref{supp:sec:results_prev_forget}, we give more experimental results to demonstrate the anti-forgetting ability of our \myeata.
    \item In Section~\ref{supp:sec:more_related_works}, we provide more discussions on related training-time robustification studies.
\end{itemize}

\section{More Implementation Details of \myeata}\label{supp:sec:more_impl}

\subsection{More Details on Datasets}\label{supp:sec:more_datasets}

Following the settings of Tent~\cite{wang2021tent} and MEMO~\cite{zhang2021memo}, we conduct experiments on three benchmark datasets for out-of-distribution generalization, \ie, CIFAR-10-C, ImageNet-C~\cite{hendrycks2019benchmarking} and ImageNet-R~\cite{hendrycks2021many}.

\textbf{CIFAR-10-C} and \textbf{ImageNet-C} consist of corrupted versions of the validation images on CIFAR-10~\cite{krizhevsky2009learning} and ImageNet~\cite{deng2009imagenet}, respectively.
The corruptions include 15 diverse types of 4 main categories
(\ie, noise, blur, weather, and digital).
Each corruption type has 5 different levels of severity.

\textbf{ImageNet-R} contains 30,000 images with various artistic renditions of 200 ImageNet classes, which are primarily collected from Flickr and filtered by Amazon MTurk annotators.

\subsection{More Experimental Protocols}
\textbf{Our \myeata.}
Following TTT~\cite{sun2020test} and Tent~\cite{wang2021tent}, we use ResNet-26 and ResNet-50 for CIFAR-10 and ImageNet experiments, respectively. The models are trained on the original CIFAR-10 or ImageNet training set and then tested on clean or the aforementioned OOD test sets.
For fair comparison, the parameters of ResNet-50 are directly obtained from the \textit{torchvision}\footnote{\href{https://github.com/pytorch/vision}{https://github.com/pytorch/vision}} library. ResNet-26 is trained via the official code of TTT by the same hyper-parameters, replacing the group norm with the batch norm, and removing the rotation head. 
For test time adaptation, we use SGD as the update rule, with a momentum of 0.9, batch size of 64, and learning rate of 0.005/0.00025 for CIFAR-10/ImageNet (following Tent and MEMO). The entropy constant $E_0$ in Eqn.~(\ref{eq:lambda_1st}) is set to $0.4\times\ln C$, where $C$ is number of task classes. The $\epsilon$ in Eqn.~(\ref{eq:lambda_2nd}) is set to 0.4/0.05 for CIFAR-10/ImageNet. The trade-off parameter $\beta$ in Eqn.~(\ref{eq:overall_loss}) is set to 1/2,000 for CIFAR-10/ImageNet to make two losses have the similar magnitude. We use 2,000 samples to calculating $\omega(\theta_i)$ in Eqn.~(\ref{eq:fisher_information}).

\textbf{Compared Methods.} For TTA~\cite{ashukha2020pitfalls}, BN adaptation~\cite{schneider2020improving} and MEMO~\cite{zhang2021memo}, the hyper-parameters follow their original papers or MEMO. Specifically, the augmentation size of TTA~\cite{ashukha2020pitfalls} is set to 32 and 64 for CIFAR-10 and ImageNet, respectively. For BN adaptation~\cite{schneider2020improving}, both the batch size $B$ and prior strength $N$ are set to 256. The hyper-parameter settings of MEMO~\cite{zhang2021memo} can be found in their original paper. For Tent~\cite{wang2021tent}, we use SGD as the update rule with a momentum of 0.9. The batch size is 64 for both ImageNet and CIFAR-10 experiments. The learning rate is set to 0.00025 and 0.005 ImageNet and CIFAR-10, respectively. Note that the hyper-parameters of Tent are totally the same as our \myeata for a fair comparison. For TTT~\cite{sun2020test}, we strictly follow their original settings except for the augmentation size at test time for ImageNet experiments. According to TTT's implementation, the augmentation size is set to 64, which, however, is very time-consuming (\eg, about 12 GPU hours on a single Tesla V100 GPU on ImageNet-C with a specific corruption type and severity level). In our implementation, we decrease this augmentation size to 20, which has only a slight performance difference compared with augmentation size 64. For example, the performances of TTT on ImageNet-C (Gaussian noise, severity level 5) with ResNet-18  are 26.2\% (64) \vs~26.0\% (20). We recommend the original papers of the above methods to readers for more implementation details.

\section{More Results on Out-of-distribution Performance and Efficiency}\label{supp:sec:results_perf_efc}

In Table~\ref{tab:imagenet-c-levels}, we provide more results to compare our \myeta and \myeata with state-of-the-art methods on ImageNet-C with the severity level 1-4.
Our \myeta and \myeata constantly outperform the state-of-the-art methods (\eg, TTA, MEMO, and Tent) in most image corruption types of various severity levels.
The performance gain mainly comes from the removal of the high-entropy test samples, since these samples may contribute unreliable and harmful gradients during test-time adaptation.

\begin{figure*}[t]
\centering
\includegraphics[width=1.\linewidth]{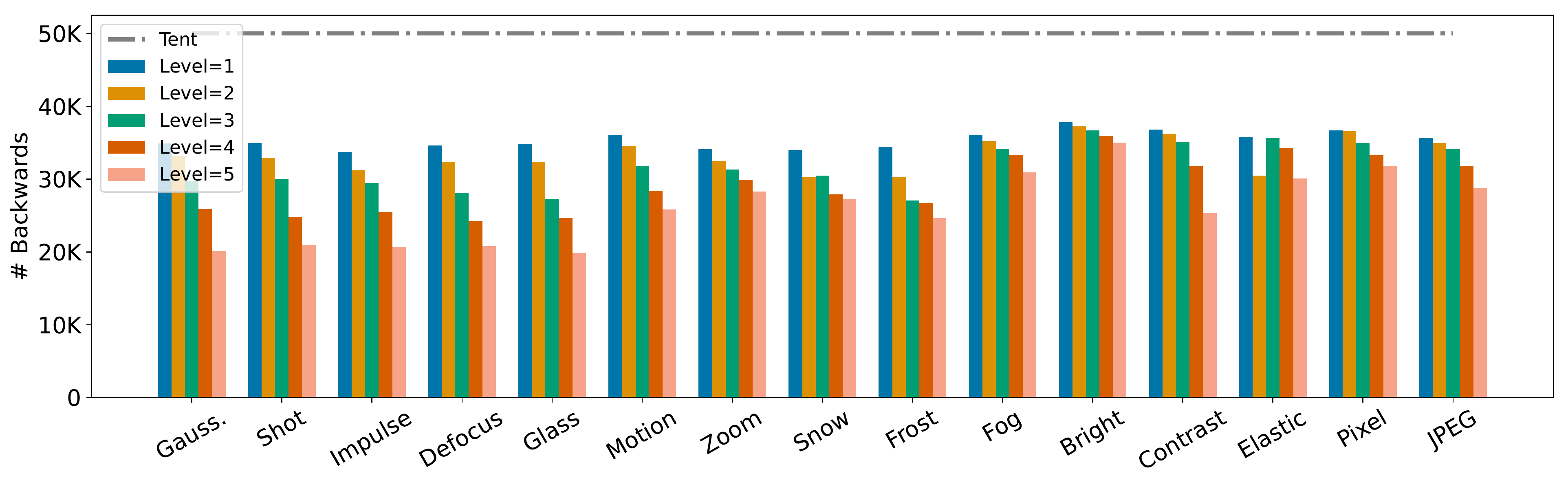}
\vspace{-0.3in}
\caption{Comparison between \myeta and Tent in terms of the number of backward propagation on ImageNet-C with different corruption types and severity levels.}
\label{fig:imageC-num-backwards2}
\end{figure*}

In Figure~\ref{fig:imageC-num-backwards2}, we show the number of backward propagation of our \myeta on ImageNet-C with different corruption types and severity levels. Across various corruption types, our \myeta shows great superiority over existing methods in terms of adaptation efficiency. Compared with MEMO (50,000$\times$64) and Tent (50,000), our \myeta only requires 31,741 backward passes (averaged over 15 corruption types) when the severity level is set to 3. The reason is that we exclude some unreliable and redundant test samples out of test-time optimization. In this case, we only need to perform backward computation on those remaining test samples, leading to improved efficiency.

\textbf{Comparison with Tent using Different Learning Rates.}
In our sample-adaptive weight $S{(\bx)}$ in Eqn.~(\ref{eq:lambda_2nd}), each test sample has a specific weight $S{(\bx)}$ and the value of $S{(\bx)}$ is always larger than 1. In this sense, training with sample-adaptive weight $S{(\bx)}$ indeed has the same effect as training with larger learning rates. Therefore, we compare our \myeata with the baseline (Tent) using different learning rates.
We increase the learning rate from $2.5\times10^{-4}$ (which is the default of Tent) to $25.0\times10^{-4}$ and report results in Table~\ref{tab:eta_vs_tent_diff_lrs}.

With the learning rate increasing from $2.5\times10^{-4}$ to $10.0\times10^{-4}$, the error of Tent decreases from 45.3\% to 43.9\%, indicating that a larger learning rate may enhance the performance in some cases. However, when the learning rate becomes larger to $20.0\times10^{-4}$, the performance of Tent degrades.
More critically, our \myeata method outperforms Tent with varying learning rates.
These results verify that simply enlarging the learning rate is not able to achieve competitive performance with our proposed sample-adaptive adaptation method, demonstrating our superiority.

\begin{table}[h]
    \vspace{-0.1in}
    \centering
    \caption{Comparison with Tent under different learning rates ($\times10^{-4}$) on ImageNet-C (Gaussian noise) regarding Error (\%). 
    }
    \label{tab:eta_vs_tent_diff_lrs}
\newcommand{\tabincell}[2]{\begin{tabular}{@{}#1@{}}#2\end{tabular}}
 \begin{center}
 \begin{threeparttable}
    \resizebox{0.55\linewidth}{!}{
 	\begin{tabular}{|l|c|cccc|}
 	\multicolumn{1}{c}{} & \multicolumn{1}{c}{\myeata (ours)} & \multicolumn{4}{c}{Tent~\cite{wang2021tent}} \\
 	\toprule
 	 Severity & $lr=2.5$ & $lr=2.5$ & $lr=10.0$ & $lr=20.0$ & $lr=25.0$ \\
 	\midrule
        Level 3   & \textbf{42.6} &	45.3 &	43.9 &	44.4 &	45.1  \\ 
        Level 5   & \textbf{65.0} &	71.6 &	72.2 &	83.6 &	87.1  \\ 
    \bottomrule
	\end{tabular}
	}
	 \end{threeparttable}
	 \end{center}
	 \vspace{-0.15in}
\end{table}

\textbf{\myeata under Different Random Orders.} \camera{Table \ref{tab:mean_and_std} records \myeata's performance (mean \& stdev.) on randomly shuffled test samples with 10 different random seeds (from 2020 to 2029). From the results, \myeata performs consistently across different random orders, showing the stability of \myeata.}

\begin{table}[h]
    \caption{The mean and stdev. of corruption accuracy (\%) of \myeata over 10 random orders, on ImageNet-C (level 5) with ResNet-50.}
    \label{tab:mean_and_std}
\newcommand{\tabincell}[2]{\begin{tabular}{@{}#1@{}}#2\end{tabular}}
 \begin{center}
 \begin{threeparttable}
    \resizebox{0.7\linewidth}{!}{
 	\begin{tabular}{|ccc|cccc|c|}
    \toprule
 	 Gauss. & Shot & Impul. & Defoc. & Glass & Motion & Zoom & Avg. \\
 	\midrule
     34.9$_{\pm 0.2}$ 	&36.9$_{\pm 0.1}$ 	&35.8$_{\pm 0.2}$ 	&33.6$_{\pm 0.3}$ 	&33.3$_{\pm 0.2}$ 	&47.2$_{\pm 0.3}$ 	&52.7$_{\pm 0.1}$  &35.8$_{\pm 0.2}$ \\
    \bottomrule
	\end{tabular}
	}
	 \end{threeparttable}
	 \end{center}
\end{table}

\section{More Results on Prevent Forgetting}\label{supp:sec:results_prev_forget}

\begin{figure}[t]
\centering     
\subfigure{\label{fig:imageC-forgetting-level-1}\includegraphics[width=1.0\columnwidth]{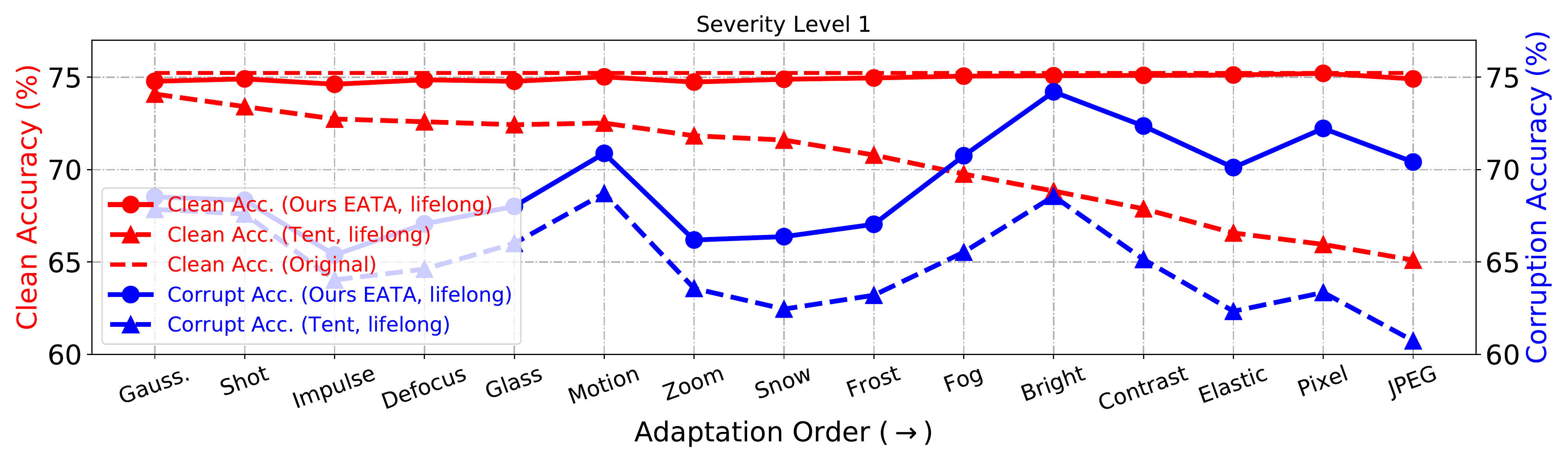}}
\subfigure{\label{fig:imageC-forgetting-level-2}\includegraphics[width=1.0\columnwidth]{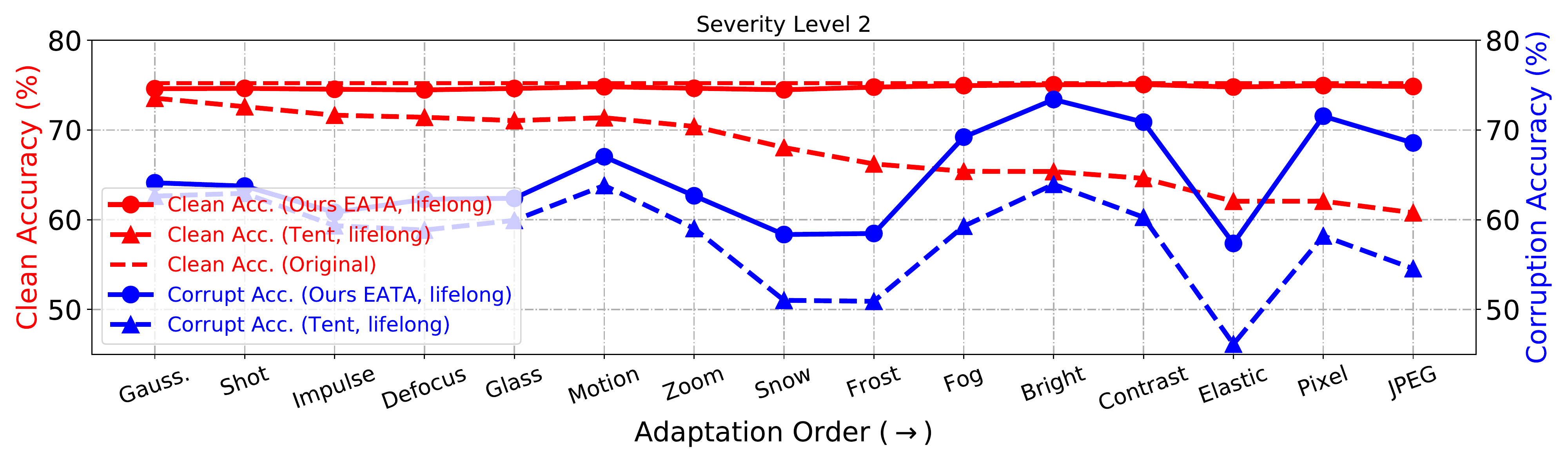}}
\subfigure{\label{fig:imageC-forgetting-level-3}\includegraphics[width=1.0\columnwidth]{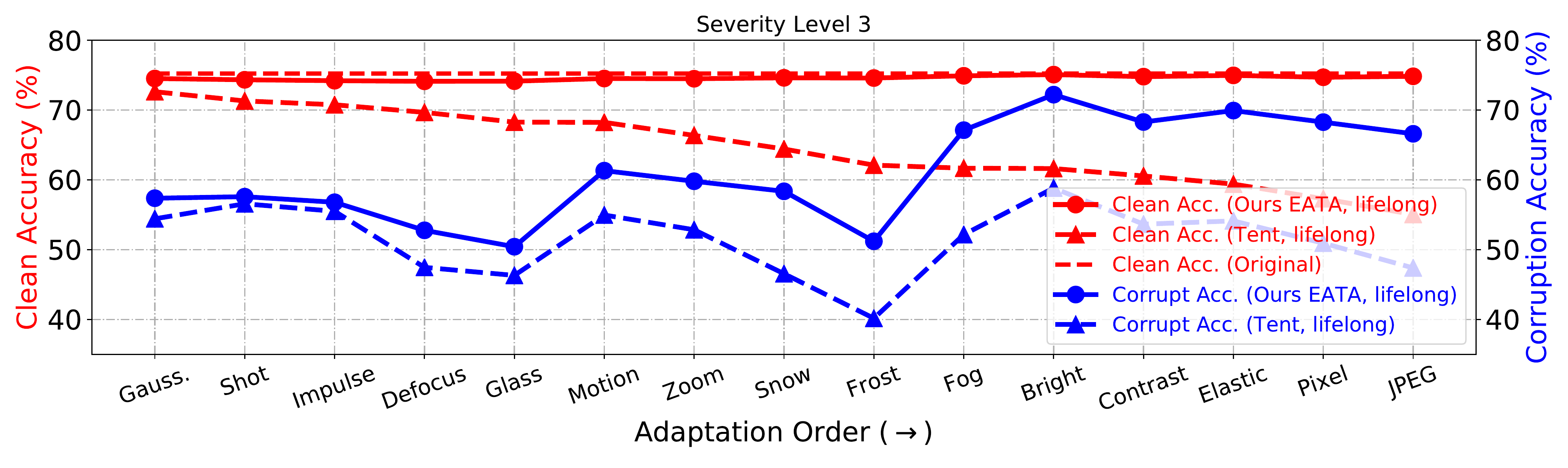}}
\subfigure{\label{fig:imageC-forgetting-level-4}\includegraphics[width=1.0\columnwidth]{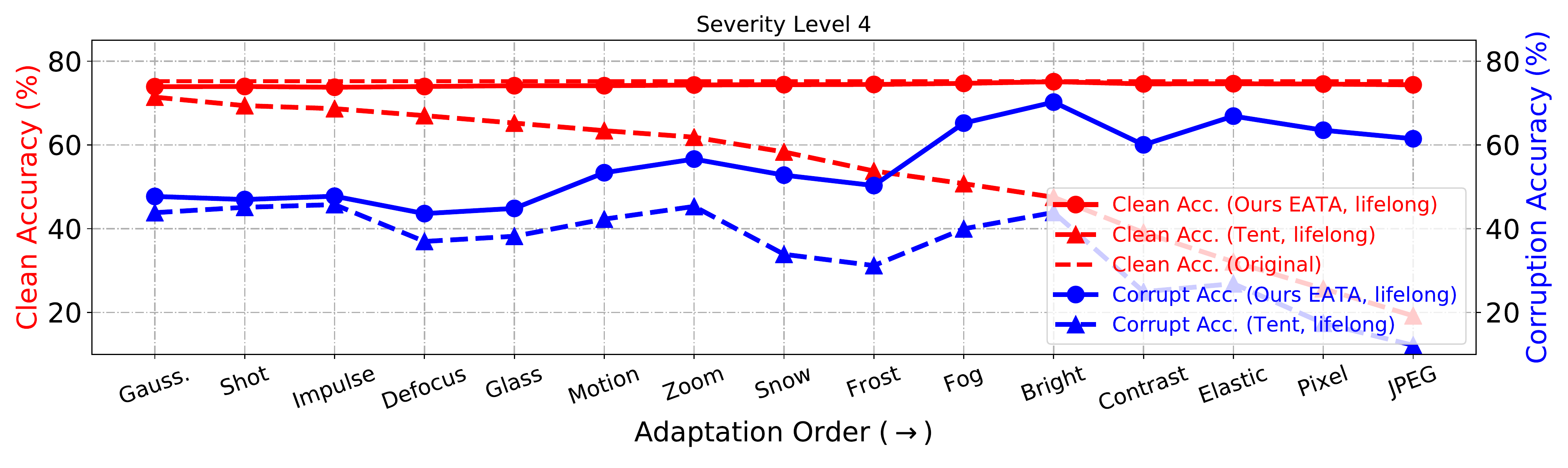}}
\caption{Comparison of prevent forgetting on ImageNet-C (severity levels 1-4) with ResNet-50. We record the OOD corruption accuracy on each corrupted test set and the associated ID clean accuracy (after OOD adaptation).  The model performs lifelong adaptation, in which the model parameters will never be reset.}
\label{fig:imageC-forgetting-levels}
\end{figure}

\begin{figure}[t]
\centering     
\subfigure{\label{fig:imageC-forgetting-level-1-each-reset}\includegraphics[width=1.0\columnwidth]{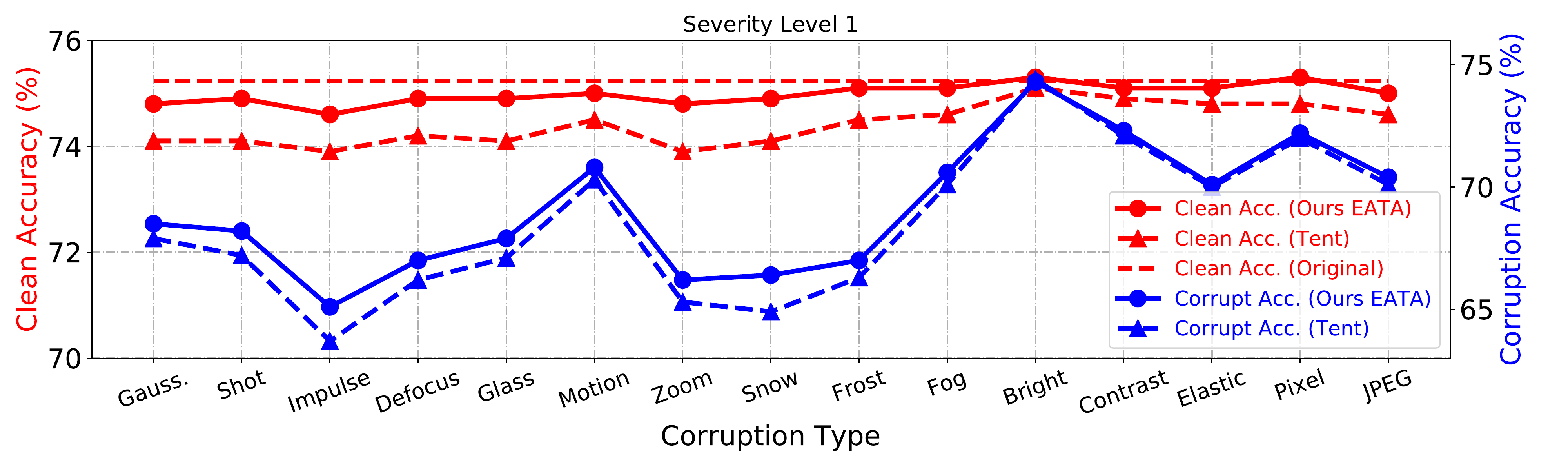}}
\subfigure{\label{fig:imageC-forgetting-level-2-each-reset}\includegraphics[width=1.0\columnwidth]{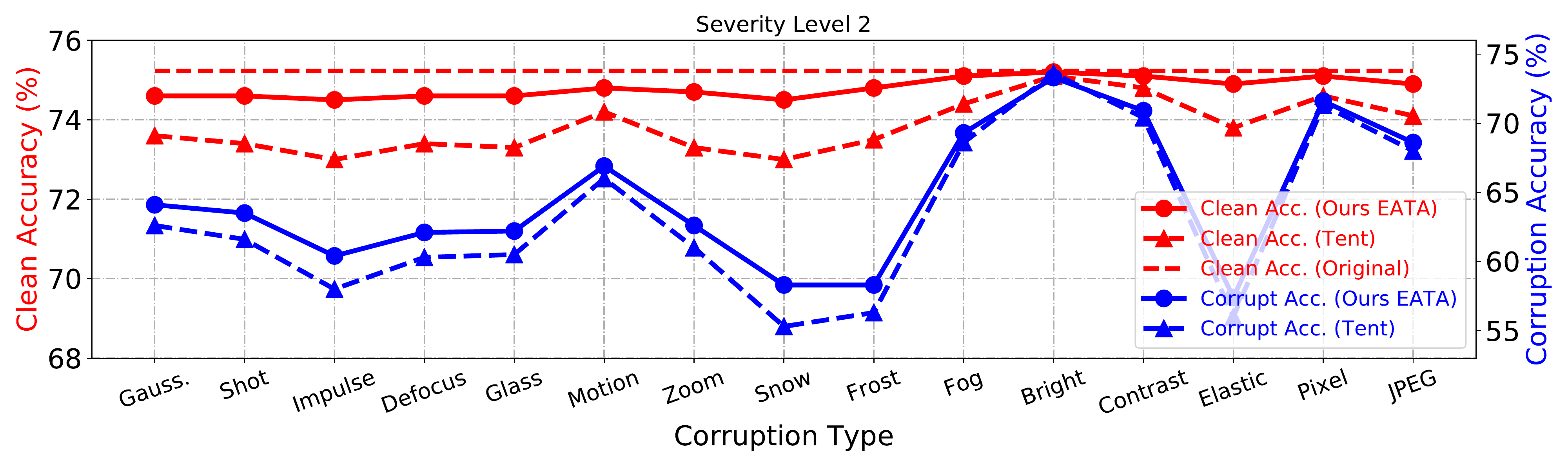}}
\subfigure{\label{fig:imageC-forgetting-level-3-each-reset}\includegraphics[width=1.0\columnwidth]{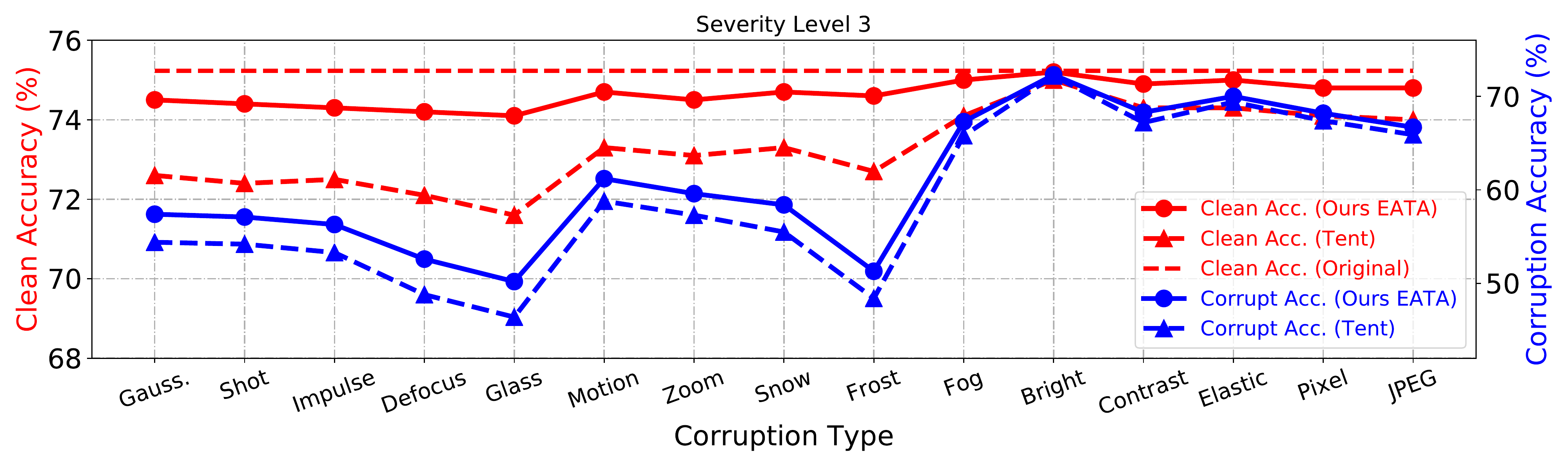}}
\subfigure{\label{fig:imageC-forgetting-level-4-each-reset}\includegraphics[width=1.0\columnwidth]{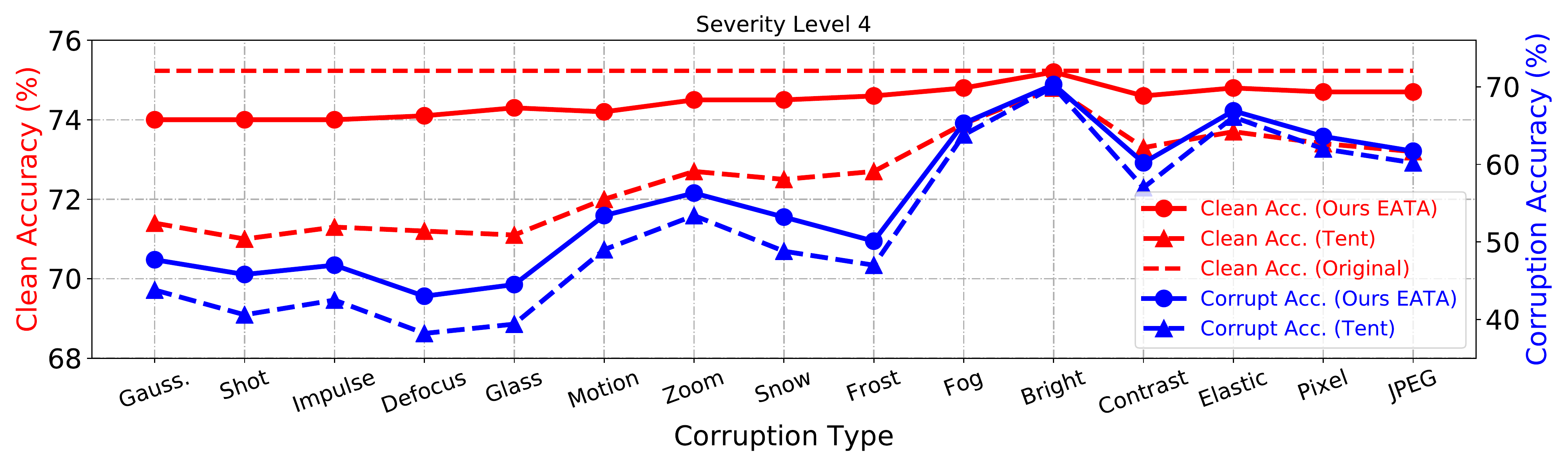}}
\caption{Comparisons of prevent forgetting on ImageNet-C (severity levels 1-4) with ResNet-50. We record the OOD corruption accuracy on each corrupted test set and the associated ID clean accuracy (after OOD adaptation).  The model parameters of both Tent and our \myeata are reset before adapting to a new corruption type.}
\label{fig:imageC-forgetting-levels-each-reset}
\end{figure}

In this section, we provide more results to demonstrate the effectiveness of our \myeata in preventing forgetting. We report the comparison results of \myeata (lifelong) \vs~Tent (lifelong) and \myeata \vs~Tent in Figures~\ref{fig:imageC-forgetting-levels} and~\ref{fig:imageC-forgetting-levels-each-reset}, respectively.  In the lifelong adaptation scenario, Tent suffers more severe ID performance degradation than that of reset adaptation (\ie, Figure~\ref{fig:imageC-forgetting-levels-each-reset}), showing that the more optimization steps, the more severe forgetting. Moreover, with the increase of the severity level, the ID clean accuracy degradation of Tent increases accordingly. This result indicates that the OOD adaptation with more severe distribution shifts will result in more severe forgetting. In contrast, our methods achieve higher OOD corruption accuracy and meanwhile maintain the ID clean accuracy (competitive to the original accuracy that tested before any OOD adaptation) in both two adaptation scenarios (reset and lifelong). These results are consistent with that in the main paper and further demonstrate the effectiveness of our proposed anti-forgetting weight regularization.

\section{More Discussions on Related Training-Time Robustification} \label{supp:sec:more_related_works}

To defend against distribution shifts, many prior studies seek to enlarge the training data distribution to enable it to cover the possible shift that might be encountered at test time, such as adversarial training strategies~\cite{wong2020fast,rusak2020simple,madaan2021learning}, various data augmentation techniques~\cite{lim2019fast,hendrycks2019augmix,li2021feature,hendrycks2021many,yao2022improving} and searching/enhancing sub-networks of a deep model~\cite{zhang2021can,guo2022improving}. However, it is hard to anticipate all possible test shifts at training time. In contrast, we seek to conquer this test distribution shift by directly learning from test data.

\clearpage

\begin{table*}[t!]
    \caption{Comparisons with state-of-the-art methods on ImageNet-C with the severity levels 1-4 regarding \textbf{Error (\%)}.
    ``GN" and ``BN" denote group normalization and batch normalization, respectively. ``JT" denotes the model is jointly trained via supervised cross-entropy loss and rotation prediction loss. The \textbf{bold} number indicates the best result and the \underline{underlined} number indicates the second best result.
    }
    \label{tab:imagenet-c-levels}
\newcommand{\tabincell}[2]{\begin{tabular}{@{}#1@{}}#2\end{tabular}}
 \begin{center}
 \begin{threeparttable}
    \resizebox{1.0\linewidth}{!}{
 	\begin{tabular}{|l|ccc|cccc|cccc|cccc|cc|}
 	\multicolumn{1}{c}{} & \multicolumn{3}{c}{Noise} & \multicolumn{4}{c}{Blur} & \multicolumn{4}{c}{Weather} & \multicolumn{4}{c}{Digital} & \multicolumn{2}{c}{Average} \\
 	\toprule
 	 Severity level=1 & Gauss. & Shot & Impul. & Defoc. & Glass & Motion & Zoom & Snow & Frost & Fog & Brit. & Contr. & Elastic & Pixel & JPEG & \#Forwards & \#Backwards \\
 	\midrule
        R-50 (GN)+JT   & 40.9 & 43.4 & 52.4 & 47.4 & 48.2 & 39.9 & 56.2 & 48.4 & 45.5 & 45.2 & 31.6 & 35.5 & 36.7 & 36.7 & 39.2 & 50,000 & 0\\
~~$\bullet~$TTT         & 37.5 & 37.4 & 39.4 & 41.8 & 39.7 & 37.2 & 43.9 & 41.7 & 40.5 & 36.3 & 31.0 & 32.9 & 35.0 & 33.8 & 36.2 & 50,000$\times$21 & 50,000$\times$20 \\
        R-50 (BN)   & 40.6 & 42.8 & 52.0 & 40.6 & 46.0 & 35.3 & 47.5 & 45.4 & 38.7 & 38.2 & 26.0 & 35.1 & 33.4 & 35.9 & 33.8 & 50,000 & 0\\
~~$\bullet~$TTA          & 40.6 & 42.6 & 52.5 & 44.4 & 46.6 & 38.0 & 45.6 & 48.0 & 42.0 & 40.6 & 28.8 & 35.6 & 33.9 & 37.7 & 36.8 & 50,000$\times$64 & 0 \\
~~$\bullet~$BN adaptation          & 34.6 & 36.1 & 41.2 & 35.7 & 35.2 & 30.8 & 37.6 & 38.2 & 35.2 & 31.3 & \textbf{25.4} & 28.5 & 30.8 & 28.7 & 30.5 & 50,000 & 0 \\
~~$\bullet~$MEMO        & 36.9 & 39.5 & 46.3 & 38.2 & 41.1 & 32.8 & 42.7 & 40.4 & 36.8 & 35.3 & 25.8 & 31.3 & 31.2 & 32.4 & 32.9 & 50,000$\times$65 & 50,000$\times$64 \\
~~$\bullet~$Tent        & 32.2 & 32.7 & 36.2 & 33.8 & 32.8 & 29.7 & 34.6 & 35.1 & 33.6 & 29.8 & \underline{25.6} & 28.0 & 30.1 & 28.0 & 29.8 & 50,000 & 50,000 \\
~~$\bullet~$Tent (episodic)       & 36.0 & 37.8 & 41.7 & 38.8 & 38.1 & 32.0 & 39.3 & 40.5 & 37.3 & 32.6 & 26.5 & 29.8 & 31.7 & 29.9 & 32.1 & 50,000$\times$2 & 50,000 \\
~~$\bullet~$\myeta (ours) & 31.7 & \underline{31.8} & \underline{34.7} & \textbf{32.9} & 32.2 & 29.6 & 34.1 & \textbf{33.6} & 33.3 & 29.5 & 26.0 & 28.2 & 30.3 & 28.1 & 29.9 & 50,000 & 35,379 \\ 
~~$\bullet~$\myeata (ours) & \textbf{31.5} & \underline{31.8} & 34.9 & 33.0 & \underline{32.1} & \underline{29.2} & \textbf{33.8} & \textbf{33.6} & \textbf{33.0} & \underline{29.4} & 25.7 & \underline{27.7} & \textbf{29.9} & \textbf{27.8} & \textbf{29.6} & 50,000 & 34,898 \\ 
~~$\bullet~$\myeata (lifelong) & \textbf{31.5} & \textbf{31.7} & \textbf{34.6} & \textbf{32.9} & \textbf{32.0} & \textbf{29.1} & \textbf{33.8} & \textbf{33.6} & \textbf{33.0} & \textbf{29.3} & 25.8 & \textbf{27.6} & \textbf{29.9} & \textbf{27.8} & \textbf{29.6} & 50,000 & 36,675 \\ 
  	\toprule
 	 Severity level=2 & Gauss. & Shot & Impul. & Defoc. & Glass & Motion & Zoom & Snow & Frost & Fog & Brit. & Contr. & Elastic & Pixel & JPEG & \#Forwards & \#Backwards \\
 	\midrule
        R-50 (GN)+JT   & 50.1 & 55.0 & 60.0 & 55.8 & 61.6 & 49.9 & 65.5 & 68.9 & 63.1 & 52.0 & 33.5 & 38.6 & 58.0 & 39.4 & 42.8 & 50,000 & 0\\
~~$\bullet~$TTT         & 42.0 & 42.5 & 44.3 & 47.7 & 53.3 & 42.7 & 48.2 & 50.4 & 56.6 & 38.2 & 31.8 & 34.3 & 49.4 & 34.7 & 38.3 & 50,000$\times$21 & 50,000$\times$20 \\
        R-50 (BN)   & 53.8 & 57.9 & 64.2 & 48.0 & 59.6 & 45.7 & 57.4 & 68.1 & 55.9 & 44.1 & 27.6 & 41.6 & 55.2 & 35.9 & 37.5 & 50,000 & 0\\
~~$\bullet~$TTA          & 52.9 & 56.9 & 62.4 & 53.0 & 60.4 & 49.3 & 54.4 & 71.5 & 60.6 & 46.9 & 30.7 & 40.0 & 53.8 & 41.1 & 40.3 & 50,000$\times$64 & 0 \\
~~$\bullet~$BN adaptation          & 42.3 & 45.6 & 49.7 & 43.0 & 44.5 & 37.4 & 44.1 & 53.1 & 47.6 & 34.1 & \textbf{26.6} & 30.6 & 47.2 & 29.7 & 33.8 & 50,000 & 0 \\
~~$\bullet~$MEMO        & 47.1 & 51.7 & 55.9 & 44.8 & 53.4 & 41.3 & 51.8 & 58.6 & 51.6 & 40.2 & 27.2 & 35.5 & 51.0 & 33.6 & 36.2 & 50,000$\times$65 & 50,000$\times$64 \\
~~$\bullet~$Tent        & 37.2 & 38.5 & 41.8 & 39.5 & 39.4 & 34.0 & 39.2 & 44.8 & 43.6 & 31.6 & \textbf{26.6} & 29.7 & 44.1 & 29.0 & 32.1 & 50,000 & 50,000 \\
~~$\bullet~$Tent (episodic)       & 44.0 & 47.2 & 50.5 & 48.0 & 49.3 & 39.2 & 46.4 & 55.0 & 50.7 & 35.5 & 27.6 & 32.3 & 49.1 & 32.2 & 36.1 & 50,000$\times$2 & 50,000 \\
~~$\bullet~$\myeta (ours) & \textbf{35.8} & \underline{36.5} & \underline{39.5} & \textbf{37.7} & \underline{37.8} & \underline{33.1} & 37.7 & \textbf{41.4} & \underline{41.7} & 30.9 & 27.0 & 29.2 & 43.0 & 28.6 & 31.8 & 50,000 & 33,363 \\ 
~~$\bullet~$\myeata (ours) & \underline{35.9} & \underline{36.5} & 39.6 & 37.9 & \underline{37.8} & \underline{33.1} & \underline{37.4} & 41.7 & \underline{41.7} & \textbf{30.7} & 26.7 & \textbf{29.1} & \textbf{42.6} & \textbf{28.4} & \textbf{31.4} & 50,000 & 32,754 \\ 
~~$\bullet~$\myeata (lifelong) & \underline{35.9} & \textbf{36.2} & \textbf{39.2} & \textbf{37.7} & \textbf{37.6} & \textbf{33.0} & \textbf{37.3} & \underline{41.6} & \textbf{41.6} & \underline{30.8} & \textbf{26.6} & \textbf{29.1} & \textbf{42.6} & \underline{28.5} & \textbf{31.4} & 50,000 & 34,922 \\ 
    \toprule
 	 Severity level=3 & Gauss. & Shot & Impul. & Defoc. & Glass & Motion & Zoom & Snow & Frost & Fog & Brit. & Contr. & Elastic & Pixel & JPEG & \#Forwards & \#Backwards \\
 	\midrule
        R-50 (GN)+JT   & 64.4 & 69.3 & 66.7 & 71.5 & 83.2 & 65.8 & 71.4 & 65.5 & 73.9 & 61.3 & 36.7 & 45.2 & 44.1 & 50.2 & 45.5 & 50,000 & 0\\
~~$\bullet~$TTT         & 48.5 & 48.9 & 48.3 & 57.7 & 67.1 & 50.7 & 50.7 & 51.2 & 70.3 & 41.0 & 33.3 & 37.6 & 36.3 & 38.4 & 39.9 & 50,000$\times$21 & 50,000$\times$20 \\
        R-50 (BN)   & 72.4 & 75.0 & 74.9 & 62.0 & 83.1 & 62.3 & 64.8 & 64.8 & 67.9 & 53.4 & 30.4 & 54.0 & 44.4 & 53.8 & 40.7 & 50,000 & 0\\
~~$\bullet~$TTA          & 70.4 & 72.7 & 70.0 & 68.7 & 83.7 & 66.2 & 62.0 & 67.9 & 72.7 & 56.3 & 33.8 & 47.7 & 47.8 & 51.8 & 43.4 & 50,000$\times$64 & 0 \\
~~$\bullet~$BN adaptation          & 54.7 & 57.2 & 56.6 & 57.8 & 63.6 & 48.9 & 48.8 & 52.2 & 57.5 & 38.2 & 28.3 & 35.4 & 33.1 & 36.0 & 36.6 & 50,000 & 0 \\
~~$\bullet~$MEMO        & 62.5 & 65.7 & 63.3 & 58.8 & 76.7 & 55.9 & 58.9 & 55.6 & 62.6 & 47.9 & 29.5 & 44.0 & 41.3 & 45.0 & 39.2 & 50,000$\times$65 & 50,000$\times$64 \\
~~$\bullet~$Tent        & 45.3 & 45.9 & 46.6 & 51.1 & 53.9 & 41.2 & 42.4 & 44.4 & 51.5 & 34.2 & 27.9 & 32.9 & 30.7 & 32.6 & 34.3 & 50,000 & 50,000 \\
~~$\bullet~$Tent (episodic)        & 56.6 & 58.8 & 57.5 & 64.9 & 69.5 & 51.5 & 51.0 & 54.4 & 60.3 & 39.9 & 29.5 & 37.7 & 35.3 & 37.5 & 39.1 & 50,000$\times$2 & 50,000 \\
~~$\bullet~$\myeta (ours) & \textbf{42.4} & \textbf{42.4} & \underline{43.3} & \underline{47.3} & \textbf{49.6} & 38.9 & 40.5 & \textbf{41.2} & \textbf{48.5} & 33.1 & 28.1 & 32.0 & 30.5 & \textbf{31.7} & 33.6 & 50,000 & {31,741} \\ 
~~$\bullet~$\myeata (ours) & \underline{42.6} & 42.9 & 43.7 & 47.4 & 49.8 & \underline{38.8} & \underline{40.4} & \underline{41.6} & \underline{48.7} & \textbf{32.7} & \textbf{27.7} & \textbf{31.7} & \textbf{30.0} & 31.8 & \textbf{33.3} & 50,000 & {31,068} \\ 
~~$\bullet~$\myeata (lifelong) & \underline{42.6} & \textbf{42.4} & \textbf{43.2} & \textbf{47.2} & \textbf{49.6} & \textbf{38.7} & \textbf{40.2} & \underline{41.6} & 48.8 & \underline{32.9} & \underline{27.8} & \textbf{31.7} & \underline{30.1} & \textbf{31.7} & \underline{33.4} & 50,000 & {33,469} \\ 
 	\toprule
 	 Severity level=4 & Gauss. & Shot & Impul. & Defoc. & Glass & Motion & Zoom & Snow & Frost & Fog & Brit. & Contr. & Elastic & Pixel & JPEG & \#Forwards & \#Backwards \\
 	\midrule
        R-50 (GN)+JT   & 81.4 & 87.6 & 82.2 & 81.8 & 87.8 & 80.2 & 77.2 & 76.8 & 75.9 & 66.9 & 41.8 & 64.2 & 54.8 & 68.0 & 54.7 & 50,000 & 0\\
~~$\bullet~$TTT         & 57.5 & 58.7 & 57.1 & 64.8 & 84.9 & 60.5 & 54.4 & 57.3 & 73.1 & 43.1 & 35.5 & 46.0 & 39.6 & 44.5 & 45.1 & 50,000$\times$21 & 50,000$\times$20 \\
        R-50 (BN)   & 89.0 & 92.1 & 91.8 & 73.5 & 87.2 & 78.1 & 71.6 & 75.9 & 70.1 & 59.6 & 34.9 & 79.5 & 57.8 & 71.1 & 52.5 & 50,000 & 0\\
~~$\bullet~$TTA          & 85.9 & 89.5 & 86.1 & 79.8 & 88.4 & 80.5 & 67.9 & 79.3 & 74.4 & 62.9 & 39.3 & 65.8 & 61.0 & 66.2 & 53.0 & 50,000$\times$64 & 0 \\
~~$\bullet~$BN adaptation          & 69.3 & 74.1 & 71.2 & 70.3 & 70.5 & 63.0 & 54.8 & 62.0 & 58.9 & 41.5 & 30.8 & 50.0 & 38.7 & 44.7 & 46.0 & 50,000 & 0 \\
~~$\bullet~$MEMO        & 78.5 & 83.3 & 78.8 & 71.0 & 82.2 & 71.2 & 67.6 & 65.9 & 64.5 & 53.3 & 33.4 & 66.7 & 52.8 & 58.6 & 48.7 & 50,000$\times$65 & 50,000$\times$64 \\
~~$\bullet~$Tent        & 56.0 & 59.4 & 57.3 & 61.7 & 60.6 & 50.9 & 46.5 & 51.4 & 53.1 & 36.5 & 30.0 & 43.3 & 34.0 & 37.9 & 39.6 & 50,000 & 50,000 \\
~~$\bullet~$Tent (episodic)        & 70.7 & 75.5 & 72.0 & 77.3 & 76.5 & 66.3 & 57.2 & 64.1 & 61.8 & 43.1 & 32.0 & 56.4 & 41.4 & 46.5 & 48.9 & 50,000$\times$2 & 50,000 \\
~~$\bullet~$\myeta (ours) & \textbf{51.5} & \underline{53.4} & \underline{52.5} & 57.1 & \underline{55.4} & \textbf{46.6} & \underline{43.7} & \underline{46.9} & \textbf{49.7} & 35.1 & \underline{29.8} & \textbf{39.4} & 33.2 & \textbf{36.0} & \textbf{38.2} & 50,000 & {29,240} \\ 
~~$\bullet~$\myeata (ours) & \underline{52.3} & 54.2 & 53.0 & \underline{57.0} & 55.5 & \textbf{46.6} & \underline{43.7} & \textbf{46.8} & 49.9 & \textbf{34.7} & \textbf{29.7} & \underline{39.8} & \textbf{33.1} & \underline{36.4} & \underline{38.3} & 50,000 & {28,423} \\ 
~~$\bullet~$\myeata (lifelong) & \underline{52.3} & \textbf{53.0} & \textbf{52.2} & \textbf{56.4} & \textbf{55.1} & \textbf{46.6} & \textbf{43.4} & 47.2 & \textbf{49.7} & \underline{34.8} & \underline{29.8} & 40.0 & \textbf{33.1} & 36.5 & 38.5 & 50,000 & {31,141} \\ 
    \bottomrule
	\end{tabular}
	}
	 \end{threeparttable}
	 \end{center}
\end{table*}

\end{document}